%% file: main.tex
\documentclass[10pt,twocolumn,letterpaper]{article}

\usepackage{wacv}              

\usepackage{graphicx}
\usepackage{amsmath}
\usepackage{amssymb}
\usepackage{booktabs}
\usepackage{dsfont}
\usepackage{makecell}
\usepackage{colortbl}
\usepackage{placeins}
\usepackage{colortbl} 
\usepackage{xcolor}   
\usepackage{mathtools}
\usepackage{color}
\usepackage{dsfont}
\usepackage{graphicx}
\usepackage{booktabs}
\usepackage{multicol}
\usepackage{multirow}
\usepackage{amsmath}
\usepackage{xcolor}
\usepackage{enumitem}
\usepackage{makecell}
\usepackage{colortbl}
\usepackage[normalem]{ulem}

\usepackage{graphicx}
\usepackage{booktabs}
\usepackage{listings}
\usepackage[colorinlistoftodos,prependcaption,textsize=tiny]{todonotes}
\presetkeys{todonotes}{inline}{}
\usepackage{comment}

\usepackage[pagebackref,breaklinks,colorlinks]{hyperref}

\newcommand{\dcfaceOGNosp}{DCFace + C$_{ge}$}
\newcommand{\dcfaceOG}{\dcfaceOGNosp~}

\newcommand{\dcfaceAllNosp}{DCFace + C$_{all}$}
\newcommand{\dcfaceAll}{\dcfaceAllNosp~}

\newcommand{\dcfaceNosp}{DCFace}
\newcommand{\dcface}{\dcfaceNosp~}

\newcommand{\digifaceNosp}{DigiFace}
\newcommand{\digiface}{\digifaceNosp~}

\newcommand{\casiaNosp}{CASIA}
\newcommand{\casia}{\casiaNosp~}

\newcommand{\lfwNosp}{LFW}

\newcommand{\calfwNosp}{CALFW}

\newcommand{\agedbNosp}{AgeDB}

\newcommand{\cplfwNosp}{CPLFW}

\newcommand{\cfpfpNosp}{CFP-FP}

\newcommand{\favcidNosp}{FAVCI2D}
\newcommand{\favcid}{\favcidNosp~}

\newcommand{\genderNosp}{\texttt{gender}}
\newcommand{\originNosp}{\texttt{ethnicity}}
\newcommand{\ageNosp}{\texttt{age}}
\newcommand{\poseNosp}{\texttt{pose}}

\newcommand{\gender}{\genderNosp~}
\newcommand{\origin}{\originNosp~}
\newcommand{\age}{\ageNosp~}
\newcommand{\pose}{\poseNosp~}

\usepackage[capitalize]{cleveref}
\crefname{section}{Sec.}{Secs.}
\Crefname{section}{Section}{Sections}
\Crefname{table}{Table}{Tables}
\crefname{table}{Tab.}{Tabs.}

\usepackage{arydshln}
\usepackage{algorithm}
\usepackage{algpseudocode}
\definecolor{codegreen}{rgb}{0,0.6,0}
\definecolor{codegray}{rgb}{0.5,0.5,0.5}
\definecolor{codepurple}{rgb}{0.58,0,0.82}
\definecolor{backcolour}{rgb}{0.95,0.95,0.92}

\lstdefinestyle{mystyle}{
    backgroundcolor=\color{backcolour},   
    commentstyle=\color{codegreen},
    keywordstyle=\color{magenta},
    numberstyle=\tiny\color{codegray},
    stringstyle=\color{codepurple},
    basicstyle=\ttfamily\footnotesize,
    breakatwhitespace=false,         
    breaklines=true,                 
    captionpos=b,                    
    keepspaces=true,                 
    numbers=left,                    
    numbersep=5pt,                  
    showspaces=false,                
    showstringspaces=false,
    showtabs=false,                  
    tabsize=2
}

\def\wacvPaperID{1166} 
\def\confName{WACV}
\def\confYear{2025}
\usepackage{orcidlink}

\begin{document}

\title{Fairer Analysis and Demographically Balanced Face Generation for Fairer Face Verification}

\author{Alexandre Fournier-Montgieux$^{*1}$\orcidlink{0009-0002-7744-3179}\\
{\tt\small alexandre.fourniermontgieux@cea.fr}
\and
Michaël Soumm$^{*1}$\orcidlink{0009-0009-0435-9903}\\
{\tt\small michael.soumm@cea.fr}
\and
Adrian Popescu$^1$\orcidlink{0000-0002-8099-824X}\\
{\tt\small adrian.popescu@cea.fr}
\and
Bertrand Luvison$^1$\orcidlink{0000-0003-2475-3712}\\
{\tt\small bertrand.luvison@cea.fr}
\and 
Hervé Le Borgne$^1$\orcidlink{0000-0003-0520-8436}\\
{\tt\small  herve.le-borgne@cea.fr}
\and 
{\small $^1$Université Paris-Saclay, CEA, LIST,F-91120, Palaiseau, France}
}

\maketitle

\begin{abstract}
Face recognition and verification are two computer vision tasks whose performances have advanced with the introduction of deep representations. 
However, ethical, legal, and technical challenges due to the sensitive nature of face data and biases in real-world training datasets hinder their development. Generative AI addresses privacy by creating fictitious identities, but fairness problems remain. Using the existing DCFace SOTA framework, we introduce a new controlled generation pipeline that improves fairness. Through classical fairness metrics and a proposed in-depth statistical analysis based on logit models and ANOVA, we show that our generation pipeline improves fairness more than other bias mitigation approaches while slightly improving raw performance.
\end{abstract}

\section{Introduction}
\label{sec:intro}
Face recognition and verification technologies (FRT and FVT) have seen significant advancements in recent years, with applications ranging from security and surveillance to personal device authentication\cite{ethic, ethic2, van2020ethical}. However, the widespread adoption of face recognition models has also raised concerns about fairness and potential biases in these systems \cite{buolamwini2018gender, karkkainen2019fairface}. Studies have shown that FRT and FVT can exhibit disparities in performance across different demographic groups, particularly along the lines of gender, ethnicity, and age \cite{sarridis2023towards, wang2019racial}.
To address these fairness challenges, researchers have explored various approaches, including the development of demographically diversified datasets \cite{grother2019face, wang2019racial}, and debiasing methods \cite{robinson2020face, yang2021ramface}. In parallel, synthetic datasets, generated using computer graphics technique~\cite{bae2023digiface, wood2021fake} and generative AI models \cite{zhao2017dual,d2024improving,bae2023digiface, kim2023dcface}, offer the potential to mitigate privacy and copyright issues \cite{Exposing_ai} associated with real datasets \cite{cao2018vggface2, kemelmacher2016megaface, guo2016ms}. 
\begin{figure}
    \centering
    \includegraphics[width=0.8\linewidth]{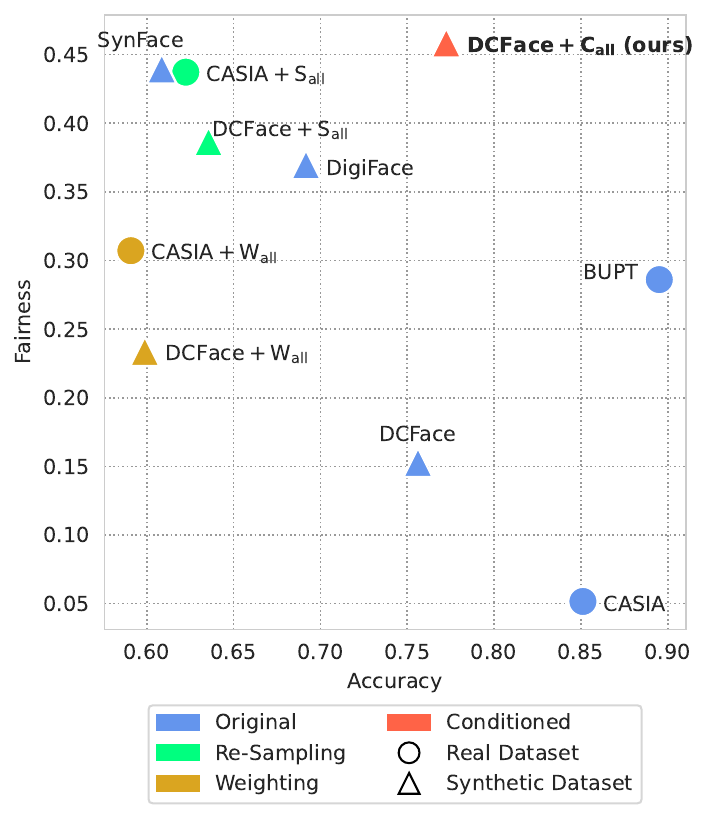}
    \caption{Comparison of the face verification fairness (equalized odds ratio) and micro-average accuracy metrics for models trained with real and synthetic images on the RFW dataset~\cite{wang2019racial}. The proposed pipeline improves the generation fairness and accuracy compared to other synthetic approaches and shows major potential for fairness mitigation compared to real sets. A performance gap in accuracy still subsists with real images. }
    \label{fig_teaser}
\end{figure}

Nonetheless, the effectiveness of synthetic datasets in improving fairness remains an open question. While existing studies highlighted the potential for generated data to reproduce or even exacerbate the biases present in real datasets \cite{perera2023analyzing}, most recent works still do not sufficiently analyze the fairness impact on models trained with their synthetic images \cite{qiu2021synface, bae2023digiface, kim2023dcface}, despite encouraging initiatives \cite{second_deandrestame_2024, Neto2023CompressedMD}. 
However, these approaches can theoretically provide greater control over data distribution and diversity.

Our first contribution, therefore, introduces a new generation control component based on the existing DCFace pipeline \cite{kim2023dcface}. The resulting approach increases the diversity of sensitive attributes such as \genderNosp, \originNosp, and \ageNosp, and also varies the \poseNosp, resulting in two new synthetic datasets \dcfaceOG, \dcfaceAll.
We compare models trained on these proposed sets with models trained using existing generation datasets, with or without bias mitigation techniques applied.

We employ a range of common metrics to measure fairness. Still, we find them insufficient for an in-depth analysis of the origins of the biases since they do not decorrelate the impacts of the considered attributes. 
We consequently introduce, as a second contribution, a new analysis approach based on logit regression models that unveils the impact of individual attributes.
Furthermore, we use an Analysis of Variance (ANOVA) to examine the relation between attributes and distance in the models' latent space. 

As highlighted in \autoref{fig_teaser}, our results demonstrate that the proposed controlled generation approach significantly improves fairness metrics while maintaining accuracy. 
The logit regression and ANOVA analyses draw coherent conclusions and reveal the effectiveness of the proposed controlled generation method in reducing attribute-based biases in both the model predictions and the latent space representations.

The code and data are released to facilitate the adoption of fairness in FRT and FVT: \url{https://github.com/afm215/FaVGen} (generation) and \url{https://github.com/MSoumm/FaVFA} (stat. analysis).

\section{Related Work}
\label{sec_related}
\begin{figure*}[!ht]
    \centering
    \includegraphics[width=\linewidth]{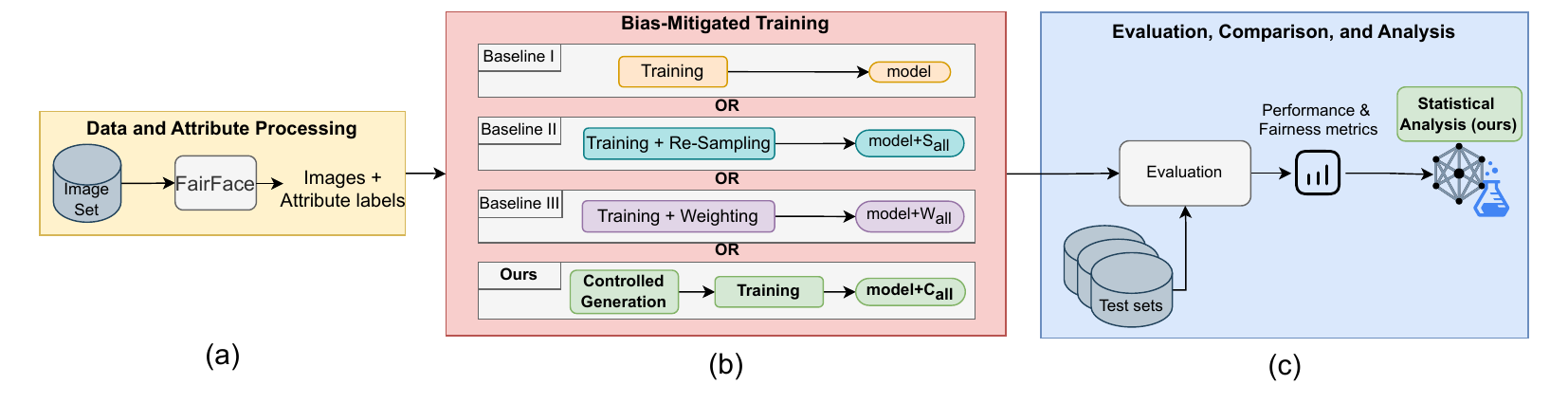}
    \caption{Global pipeline overview for training and evaluating models with the baselines and our proposed generative approach. Critical attributes are collected on image sets (a) that enable using bias mitigation techniques before or during model training (b). Models are then evaluated on FVT evaluation sets (c), and their biases are then analyzed using fairness metrics and our proposed statistical analysis. Contributions of this paper are colored green.}
    \label{fig_pipeline}
\end{figure*}
\noindent\textbf{Face verification} is a classical yet still open research topic. Following ~\cite{robinson2020face}, a model is trained to perform face recognition. Then, given a pair of images, the evaluation task is determining whether they belong to the same identity using the trained model as an embedding extractor. A threshold 
is optimized to separate and predict the positive and negative pairs. 
Following~\cite{phillips2012good,popescu2022face,wang2019racial}, we advocate for selecting hard negative images to make verification more realistic and consider datasets including difficult negatives to evaluate the models' performance. 
We also advocate for more efforts to integrate fairness in the verification evaluation process. Fairness evaluation can be improved by designing demographically-diversified verification datasets~\cite{grother2019face,popescu2022face,wang2019racial} and integrating demographic metadata in them~\cite{sarridis2023towards}. Demographic attributes balance deserves particular attention because it is required for analyzing potentially serious discrimination~\cite{sarridis2023towards, ethic}.

\noindent\textbf{Real training datasets} for face recognition are usually created by scraping a large number of images from publicly available sources~\cite{kemelmacher2016megaface,schroff2015facenet} and then cleaning them~\cite{cao2018vggface2,guo2016ms,yi2014learning} to reduce the number of unrepresentative samples. However, these datasets face several challenges. First, obtaining subjects' consent at scale is impossible, posing a serious legal challenge when collecting sensitive data such as identified faces. Second, most datasets~\cite{cao2018vggface2,guo2016ms,yi2014learning} include copyrighted photos, raising licensing issues. The lawfulness of distributing copyrighted content is a longstanding discussion that applies to other computer vision tasks~\cite{quang2021does} and was recently revived by the success of foundation models trained with very large datasets~\cite{scao2022bloom}. Third, existing large datasets exhibit demographic (gender, ethnicity, age)~\cite{popescu2022face,sarridis2023towards,wang2019racial}, face characteristics (size, make-up, hairstyle)~\cite{albiero2020does,albiero2021gendered,terhorst2021comprehensive}, and visual biases~\cite{zhao2018towards}, mostly reflecting the sampling bias affecting images datasets~\cite{fabbrizzi2022bias_visual_datasets}. These biases affect underrepresented segments~\cite{buolamwini2018gender,karkkainen2019fairface,sarridis2023towards} and should be addressed to improve fairness. These problems make the sustainable publication of real datasets very complicated, as proven by the withdrawal of most resources~\cite{cao2018vggface2,guo2016ms,kemelmacher2016megaface} following public pressure~\cite{van2020ethical}.

\noindent\textbf{Synthetic datasets} have the potential to reduce or remove privacy, copyright, and unfairness issues compared to real datasets~\cite{kim2023dcface, second_deandrestame_2024, Neto2023CompressedMD}. Computer graphics techniques are used in~\cite{bae2023digiface,wood2021fake} to render diversified face images, and strong augmentations are added to increase accuracy. Most works rely on generative AI, with~\cite{zhao2017dual} being an early example that uses dual-agent GANs to generate photorealistic faces. The authors of~\cite{qiu2021synface} identify the lack of variability of generated images as a central challenge and propose identity and domain mixup to improve synthetic datasets. Diffusion models were used very recently~\cite{kim2023dcface} to create identities and to diversify their samples based on a style bank. Synthetic datasets have the advantage of including fictitious identities, alleviating privacy and copyright issues associated with real-face datasets. 
However, privacy issues can remain regarding data replication in GANs~\cite{feng2021gans} and diffusion models~\cite{somepalli2023diffusion} but can be controlled and mitigated as shown in ~\cite{doubinsky2022prl, barattin2023attribute, doubinsky2023wasserstein}. When uncontrolled, synthetic datasets are also likely to reproduce and even exacerbate the biases of real datasets in a constrained evaluation setting~\cite{perera2023analyzing}. 

\noindent\textbf{Debiasing methods} have been proposed to mitigate biases in face verification. One approach is to adapt the verification process to demographic segments. The authors of~\cite{robinson2020face,terhorst2020comparison} propose adaptive threshold-based approaches to improve fairness. Another approach is to address ethnicity-related bias by learning disparate margins per demographic segment in the representation space~\cite{yang2021ramface,wang2021meta,wang2020mitigating} or by suppressing attribute-related information in the model~\cite{sarridis2023flac}. While technically interesting, these methods are ethically and legally problematic in practice since they assume disparate treatment of human subjects by AI-based systems. We advocate for bias mitigation directly within model training sets, which we show to have a very concrete consequence on model biases. 

\section{ Methodology }
The overall training and evaluation pipeline (\autoref{fig_pipeline}) comprises three parts: 
Part (a) regroups training sets and their attributes. These training sets may or may not be combined with bias mitigation techniques to train models (part (b)). These techniques include our proposed controlled data generation (in green). Finally, as explained in \autoref{sec_related}, these models are used in part (c) to perform FVT using the setup of \cite{robinson2020face, LFWTechUpdate }. The results obtained on FAVCI2D \cite{popescu2022face}, RFW \cite{conti2022mitigating}, and BFW \cite{bias3} are analyzed in terms of raw performance (accuracy), fairness metrics, and using the statistical approach we introduce in this paper.

Following recent face recognition work\cite{bae2023digiface, kim2023dcface}, we train models using a ResNet50 architecture~\cite{he2016deep} with a loss designed specifically for this task~\cite{kim2022adaface}.
We create face recognition models with different training sets.
We ensure comparability between these training sets by using the same structure and similar size, compatible with previous studies~\cite{bae2023digiface,qiu2021synface,yi2014learning}.
They contain 10,000 unique identities and 50 samples per identity.

\subsection{Considered Biases}

\label{subsec_AttributeAnalysis}

We balance the created datasets for four attributes: \originNosp, ~\genderNosp, ~\ageNosp, and \poseNosp.
The first three are sensitive attributes contributing directly to demographic fairness and are usually employed in the literature~\cite{bias1, bias2, bias3, bias4}.
The fourth ensures face appearance variability and augments model performance.
\origin and \gender are attributes associated with each identity.  
When unavailable in the datasets' metadata, these attributes are inferred using FairFace~\cite{karkkainen2019fairface}. 
In this case, \origin and \gender are categorical (Asian, Black, Indian, White) and binary variable (female/male).
Since they are supposed to be consistent across the images of the same identity, we mitigate the potential inference errors by averaging the FairFace outputs per identity.
Age is also inferred at the image level using FairFace.

The \pose attribute is extracted using the model introduced in~\cite{hempel20226d}. 
We use face rotation around the pitch, the yaw, and the roll axes (i.e., the rotations around the $x$, $y$, and $z$ axes) to characterize \poseNosp.

\subsection{Proposed Balanced Dataset Generation}
Our controlled approach relies on the DCFace \cite{kim2023dcface} generation pipeline. 
It applies the style of a real picture (style image) to a synthetic face picture (ID image) using a dual-conditioned diffusion model. 
DCFace combines a single ID with several style images to produce the samples representing each synthetic identity in the training set.
The identity-level attributes (\origin and \genderNosp) are, therefore, controlled by the choice of the ID image.
The picture-level attributes (\age and \poseNosp) are controlled by the choice of the style images.

We thus introduce a joint diversification process on \genderNosp, \originNosp, \ageNosp, and \pose attributes. 
We select a list of ID images generated with DDPM\cite{ddpm} whose joint \genderNosp$\times$\origin distribution is perfectly balanced.
We diversify \pose and \age by iteratively populating the less-represented age and pose categories of each identity. 
We also match the demographic segment (\genderNosp$\times$\originNosp) of ID and style images to facilitate the loss convergence process.
We implemented this matching following initial tests, which showed that convergence is not guaranteed without anything else. 
We create two versions of the balanced dataset to assess the influence of identity-level and image-level attributes.
\textbf{\dcfaceOG} uses only \genderNosp$\times$\origin,
\textbf{\dcfaceAll} considers all four attributes. 

\subsection{Training Set Baselines}

We compare \textbf{\dcfaceOG} and \textbf{\dcfaceAll} with a representative set of real and synthetic datasets:
\textbf{CASIA}~\cite{yi2014learning} - real dataset representing celebrities from the IMDB dataset. 
\textbf{BUPT}~\cite{wang2021meta}  - real dataset that is balanced for \originNosp.
Note that the full version includes more than 1M images. We subsample BUPT to match the structure of other baselines~\cite{bae2023digiface,qiu2021synface,yi2014learning}. 
\textbf{SynFace}~\cite{qiu2021synface} - synthetic dataset created with a GAN architecture using identity and domain mixup to diversify generated faces. 
\textbf{DigiFace}~\cite{bae2023digiface} - synthetic dataset created using rendering technique to obtain diversified representations of faces of each identity. 
\textbf{\dcface}~\cite{kim2023dcface} - synthetic dataset generated using the default uncontrolled pipeline of \cite{kim2023dcface}.

\begin{figure*}
    \centering
    \includegraphics[width=0.8\linewidth]{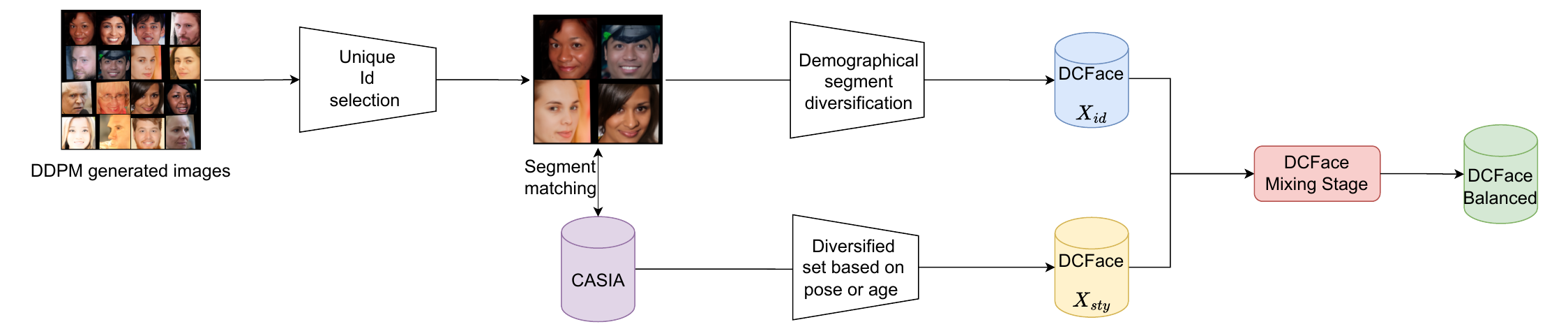}
    \caption{Detailed view of our controlled generation method.}
    \vspace{-3mm}
    \label{fig_controlled_generation}
\end{figure*}

 \subsection{Dataset Biases Analysis}

We report the attribute diversity $a$ for a dataset $\mathcal{D}$ computed as the normalized entropy applied on the frequency $p_{a_i}$ for the attribute sub-groups $a_{i \in [1,m]}$.
 \begin{equation}
 Diversity_{a}(\mathcal{D}) = - \frac{1}{\ln(N)}\sum_{i =0}^{N}  p_{a_i} \ln(p_{a_i}) 
 \end{equation}
 
 \autoref{tab_datasets} enables a data-oriented comparison of our datasets and baselines.
 It highlights the proposed pipeline's effectiveness and the need for joint attribute balancing to avoid unwanted side effects. 
 For instance, balancing ethnicity and gender alone induces a notable lack of age diversity, and our pose balancing indeed results in more pose diversity.
 For instance, only balancing on ethnicity and gender reduces age diversity and does not affect pose, while balancing for all attributes results in a better global trade-off. 

\begin{table}[!hb]

\definecolor{color1}{RGB}{255, 85, 85}    
\definecolor{color2}{RGB}{239, 96, 80}
\definecolor{color3}{RGB}{223, 107, 75}
\definecolor{color4}{RGB}{207, 118, 70}
\definecolor{color5}{RGB}{191, 129, 65}
\definecolor{color6}{RGB}{175, 140, 60}
\definecolor{color7}{RGB}{159, 151, 55}
\definecolor{color8}{RGB}{143, 162, 50}
\definecolor{color9}{RGB}{127, 173, 45}
\definecolor{color10}{RGB}{111, 184, 40}
\definecolor{color11}{RGB}{95, 195, 35}
\definecolor{color12}{RGB}{79, 206, 30}
\definecolor{color13}{RGB}{63, 217, 25}
\definecolor{color14}{RGB}{47, 228, 20}
\definecolor{color15}{RGB}{31, 239, 15}
\definecolor{color16}{RGB}{15, 250, 10}       
\resizebox{0.99\linewidth}{!}
{
\begin{tabular}[!b]{c|ccccccc}
Attribute & \casia & BUPT & \digiface & SynFace  & \dcface & \textbf{ \dcfaceOG } & \textbf{\dcfaceAllNosp}  \\
 \hline 
Gender & \cellcolor{color16} 1.00 & \cellcolor{color10} 0.93 & \cellcolor{color10} 0.93 & \cellcolor{color16} 0.99 & \cellcolor{color16} 0.99 & \cellcolor{color16} 1.00 & \cellcolor{color16} 1.00 \\
 \hline 
Ethnicity & \cellcolor{color1} 0.47 & \cellcolor{color10} 0.92 & \cellcolor{color7} 0.65 & \cellcolor{color1} 0.40 & \cellcolor{color3} 0.56 & \cellcolor{color10} 0.93 & \cellcolor{color10} 0.90 \\
 \hline 
Age & \cellcolor{color4} 0.59 & \cellcolor{color10} 0.71 & \cellcolor{color1} 0.42 & \cellcolor{color7} 0.64 & \cellcolor{color7} 0.64 & \cellcolor{color5} 0.61 & \cellcolor{color9} 0.69 \\
 \hline 
Pose & \cellcolor{color5} 0.61 & \cellcolor{color3} 0.57 & \cellcolor{color8} 0.67 & \cellcolor{color4} 0.58 & \cellcolor{color1} 0.51 & \cellcolor{color1} 0.51 & \cellcolor{color4} 0.58
\end{tabular}
}

\caption{ Inferred diversity for several training datasets.  The degree of balance is quantified by the entropy for the considered attributes across the dataset. \textbf{Datasets introduced in this paper are in bold}.}
\label{tab_datasets}

\end{table}

\subsection{Baseline Debiasing Methods}
We compare the proposed dataset bias mitigation pipeline with two classical baseline methods: resampling\cite{Oversampling1, Oversampling2, Oversampling3, Oversampling4, Undersampling1, Undersampling2, Undersampling3} and loss weighting \cite{weighting1, weighting2, weighting3}. 
We apply these common debiasing techniques on imbalanced sets (CASIA and \dcfaceNosp). 
The frequency of the considered classes determines an image's sampling probability and sample weight, which are used in resampling and weighting, respectively.
We detail these methods in the supplementary material. 
We add +S$_{ge}$ and +W$_{ge}$ to initial dataset names for resampling and loss weighting limited to \gender and \originNosp.
We add +S$_{all}$ and +W$_{all}$ when all attributes are debiased.

\section{Toward a Fairer Analysis of FVT evaluation }
\subsection{Evaluation Sets and Protocol}
We use RFW \cite{conti2022mitigating}, FAVCI2D \cite{popescu2022face}, and BFW~\cite{robinson2020face} in our fairness analysis.
We selected the two first face verification datasets because they have sufficient identities per demographic segment for rigorous analysis, and the third one because of its balancing.
We provide additional accuracy-oriented results using classical datasets, such as LFW~\cite{LFWTechUpdate}, AgeDB~\cite{moschoglou2017agedb}, and CPLFW~\cite{zheng2018cplfw} in the supplementary material. 
These datasets are either too small or demographically imbalanced to enable robust fairness in assessment. 

\begin{figure}[tb]
    \centering
\includegraphics[width=\linewidth]{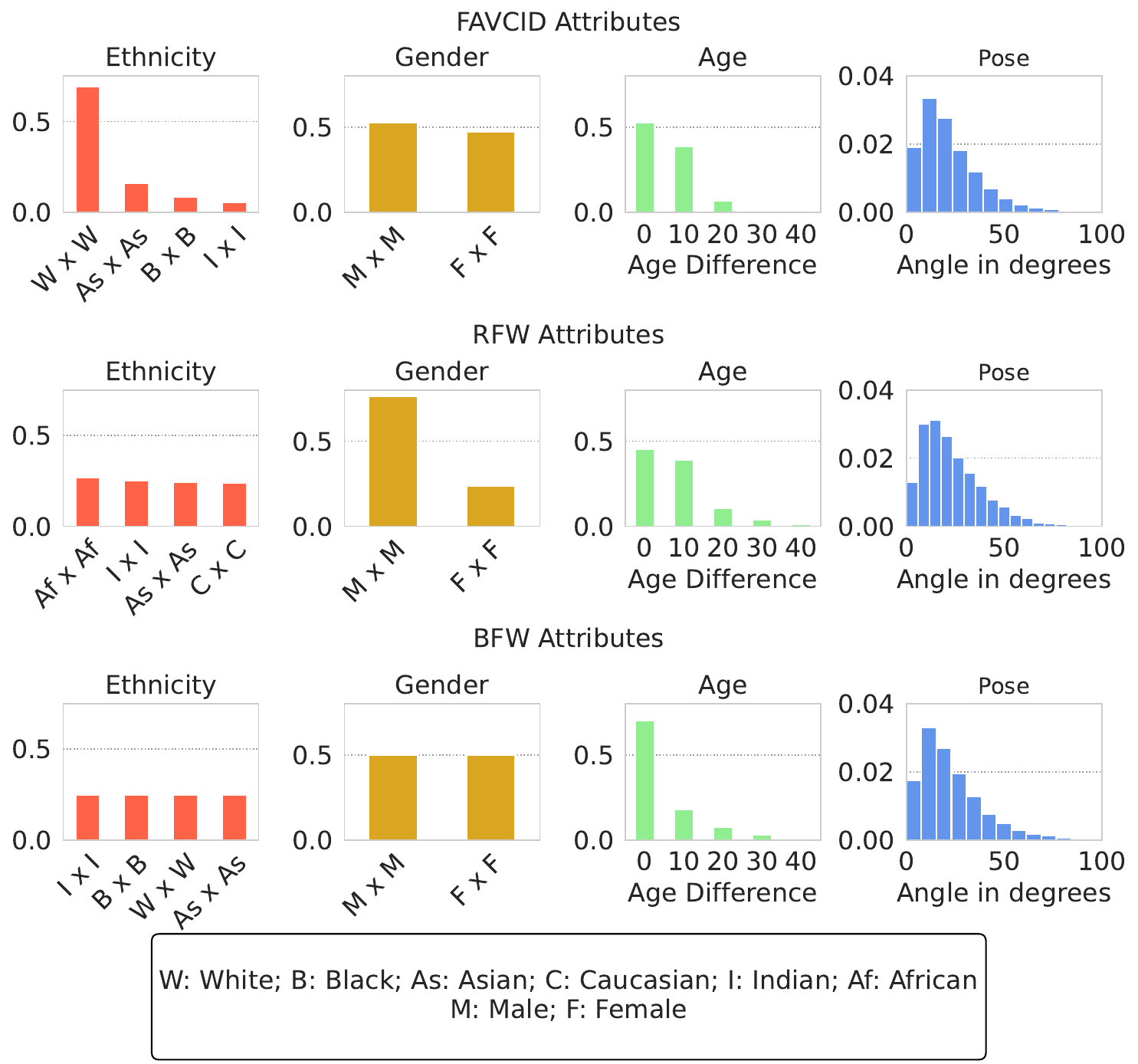}
    \caption{Attribute analysis of the evaluation datasets. Attributes are generated using FairFace \cite{karkkainen2019fairface}, except for the \gender of BFW, and \origin for RFW and BFW included with the datasets.}
    \vspace{-3mm}
    \label{fig_attributes}
\end{figure}

Similar to training datasets, we extract FairFace attributes whenever they are not provided. 
For RFW and BFW, we use the included \origin (as well as \gender for BFW) attribute since the datasets are already balanced for it.
\autoref{fig_attributes} presents a brief description of the pair attributes in the RFW, FAVCI2D, and BFW datasets. While the three datasets have similar balancing on \age and \pose attributes, they exhibit different characteristics in terms of gender and ethnicity distributions. FAVCI2D has a relatively balanced gender distribution but a skewed ethnicity distribution, with the \texttt{White} ethnicity being the most prevalent. In contrast, RFW has a more balanced representation of ethnicity, with a uniform distribution, like BFW, across \texttt{African}, \texttt{Indian}, \texttt{Asian}, and \texttt{Caucasian} ethnicities, but is unbalanced in terms of \genderNosp, unlike BFW. Despite being balanced, BFW has very few identities that might introduce singularity. This can explain the surprising behavior of fairness metrics on this dataset (e.g. CASIA being fair for most metrics). These differences allow for a comprehensive evaluation of face verification models' fairness and performance across diverse demographic groups, assessing how well the models handle gender and ethnicity variations and identifying potential biases arising from imbalanced training data.

\subsection{Fairness and Performance Metrics}

To evaluate face recognition performance, we consider the following metrics:
\textbf{Micro-Average Accuracy}\cite{performancemetrics1} is commonly used for evaluating the overall performance of a face recognition model. 
It is particularly useful when dealing with unbalanced data, as it gives equal weight to each dataset segment, regardless of the group size. 
Consequently, the overall accuracy is not biased toward the majority group. 
\textbf{True Match Rate (TMR)\footnote{is equivalent to $1-$FNMR}, or TPR}, measures the proportion of actual positive cases that are correctly identified. \textbf{False Match Rate (FMR), or FPR}, measures the proportion of negative cases incorrectly identified as positive by the face recognition model.
We follow existing face recognition literature~\cite{ethic, van2020ethical} and consider FMR as a more critical metric compared to TMR.

To evaluate face recognition fairness, we consider the following metrics:
\textbf{Degree of Bias (DoB}) \cite{gong2020joint_debiaising} is the standard deviation of accuracy across different subgroups, which is higher when the performance varies a lot w.r.t each subgroup. However, datasets with low accuracy tend to have a smaller overall variance inherently. 
Moreover, DoB does not allow for fine-grained error analysis, which is central to understanding performance variations in our case. 
\textbf{Demographic Parity Difference (DPD)} and \textbf{Demographic Parity Ratio (DPR)} \cite{equlizeodddemographocparity, demographicparity} require that the probability for individuals to receive a positive outcome should be the same across all demographic groups. DPD is the absolute difference between the highest and lowest probability across all subgroups, whereas DPR is the ratio between the lowest and highest. The closer the DPD is to zero and the closer the DPR is to one, the fairer the results are. 
\textbf{Equalized Odds Difference (EOD)} and \textbf{Equalized Odds Ratio (EOR)} \cite{equlizeodddemographocparity,equlizedodd}  require that the face recognition model's  TMR and FMR are independent of the demographic groups, thus ensuring consistent accuracy across groups. 
EOD is calculated as the maximum absolute difference between the TMRs or FMRs across groups. EOR is the minimum between the ratios of the TMRs and FMRs across groups. The closer the EOD is to zero
and the closer the EOR, the fairer the results are.

\input{result_table}

\subsection{Proposed Statistical Analysis Approach}
Our statistical analysis pipeline comprises logit regression~\cite{angrist2009mostly} and Analysis of Variance (ANOVA)~\cite{gareth2013introduction}.
These methods provide complementary insights into the impact of the studied attributes on fairness.

Logit regression \cite{angrist2009mostly} models the relation between attributes and the binary outcome of a model. It is a generalized linear model that estimates the probability of a binary outcome based on one or more independent variables using:
\begin{equation}
  \ln \dfrac{\mathbb{P}[y=1|X]}{\mathbb{P}[y=0|X]} = \beta_0 + \beta_1 X_1 + ... +\beta_k X_k 
\end{equation}
where $y$ is a target binary variable to explain, $X_1, ...X_k$ are the $k$ explanatory variables; and $\beta_0, ..., \beta_k$ are the fitted coefficients. In the fairness analysis context, for an image pair given as input to a face verification algorithm, the binary outcome represents the quantity $\mathds{1}(y_{pred}=y_{true})$. When properly fitted, the logit regression coefficients represent the impact on the binary outcome for a unit change in the corresponding attribute, holding other attributes constant.

ANOVA~\cite{gareth2013introduction} is used to determine whether significant differences exist among the means of multiple groups. In fairness analysis, ANOVA can be applied to a continuous variable, such as the distance between face representations in the latent space, to measure the importance of each attribute in explaining the observed variations. By treating the latent space distance as the dependent variable and the attributes as independent variables or factors, ANOVA can partition the total variance in the distances into components attributable to each attribute. Additionally, a quantity named $\eta^2$ can be computed for each variable and used to represent the variance the variable explains. 

ANOVA identifies the overall importance of attributes in explaining variations in the latent space, while logit regression quantifies the specific impact of attributes on binary identification outcomes.
Section~\ref{sec_res} presents the detailed application of these methods to the datasets and fairness metrics, along with the interpretation of the results. 

\section{Results and Analysis}
\label{sec_res}
We report and discuss fairness metrics on both \favcid and RFW sets. Statistical and ANOVA analysis is performed on RFW and is reported for \favcid in the supplementary material. 
\begin{figure*}[h]
    \centering
    \includegraphics[width=\linewidth]{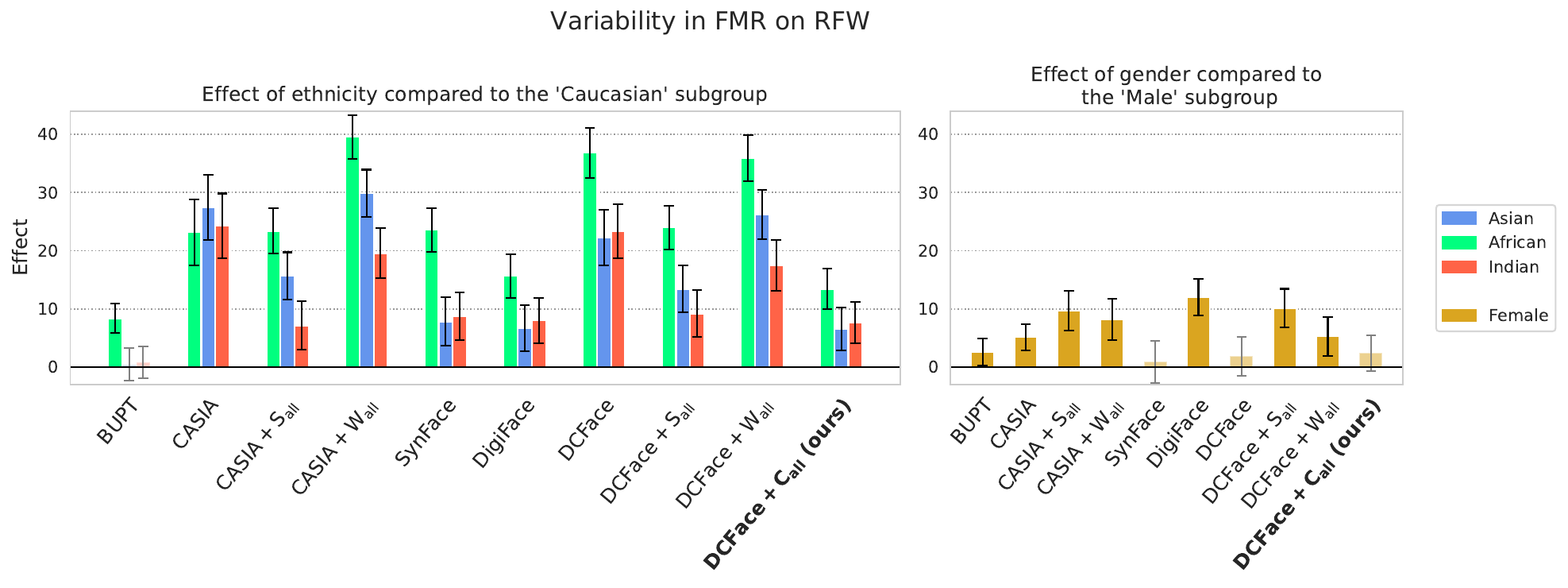}
    \caption{Marginal effect on FMR (lower is better) for each method compared to the unprotected group. Example: "When using \casia, on average and other things being equal, two people from the \texttt{African} subgroup are 22\% more likely to be wrongly misidentified than two people from the \texttt{Caucasian} subgroup". Non-significant effects are shown in transparency. Our controlled generation reduces biases of \dcface more effectively than other bias mitigation techniques.}
    \vspace{-3mm}
    \label{fig_FPR}
\end{figure*}

\subsection{Performance \& Fairness Comparison}

We report the fairness metrics and micro-average accuracy for all training approaches on RFW, FAVCI2D, and BFW, both for real and synthetic datasets (\autoref{tab_results}). 

Among the real datasets, the model trained on BUPT achieves higher accuracy than models trained on CASIA on RFW, FAVCI2D and BFW. 
BUPT also gets the best fairness metrics on \favcidNosp and BFW but surprisingly, on RFW, it shows a mitigated behavior, being first in terms of EOD only. Overall on RFW, CASIA+S$_{ge}$ shows the best behavior in terms of fairness (DPR, DPD, EOD), at the cost of 5.2 points of accuracy compared to the original CASIA set. This surprising behavior is not noticed with our in-depth analysis (especially in \autoref{fig_FPR}), which draws other conclusions for BUPT model sensitivity, advocating for the usefulness of our analysis approach. 
Findings are similar for BFW, with CASIA fairness being close to BUPT's. 
This surprising behavior might come from the few identities that compose the dataset.
BUPT accuracy is also better than CASIA's, with some variability between the three tested verification datasets.
This is probably the result of a different degree of shift between train and verification data. 

Among synthetic datasets, the proposed \dcfaceOG and \dcfaceAll show the most promising results across the evaluation sets. 
These balanced variants improve fairness compared to DCFace, the original generation pipeline they build upon.
The fairness gains are large for DPD, EOD, DPR, and EOR and less important for DoB. 
The differences between \dcfaceAll and \dcfaceOG are small for most fairness metrics, but \dcfaceAll provides a mild accuracy gain.
The results demonstrate that the proposed balancing pipeline, particularly DCFace + C$_{all}$, substantially improves fairness metrics across different verification datasets.
Importantly, a small accuracy gain compared with the original DCFace dataset is also observed, along with fairness improvement.
The models trained with balanced datasets probably benefit from a smaller shift between training and verification datasets, reflected in the micro-average accuracy measured during evaluation. 
Similar results are obtained in terms of raw accuracy and are reported in the supplementary on five additional verification sets used in prior works 
~\cite{bae2023digiface,kim2023dcface}.

\subsection{Logit Model for Bias Quantification}
To quantify the biases in face recognition outcomes more precisely, we employ a logit model that estimates the impact of person attributes on face verification model predictions. 
Hence, we examine the relationship between the studied attribute and the face recognition system's performance in terms of FMR and TMR. The two logit regressions are:
\begin{align*}
    \text{(TMR)}\ \mathds{1}(\hat{y}=1|y=1) = \sigma(&\beta_0 + \beta_1 \cdot\texttt{ethnicity} \\+ &\beta_2 \cdot\texttt{gender}+ \beta_3 \cdot\texttt{age}\\+ &\beta_4 \cdot\texttt{pose})\\
    \text{(FMR)} \ \mathds{1}(\hat{y}=1|y=0) = \sigma(&\beta_0 + \beta_1 \cdot\texttt{ethnicity} \\+ &\beta_2 \cdot\texttt{gender}+ \beta_3 \cdot\texttt{age}\\+ &\beta_4 \cdot\texttt{pose})
\end{align*}
where $\hat{y}$ is the prediction of the model; $y$ is the ground-truth label of the pair; $\sigma$ is the sigmoid function; \texttt{ethnicity} and \texttt{gender} are categorical variables implemented with the dummy variable coding \cite{hardy1993regression}; \texttt{age} and \texttt{pose} are handled as continuous variables.

The logit model coefficients $\beta_k$ represent the change in the log odds of the binary outcome (e.g., false positive or true positive) for a unit change in the corresponding attribute, holding other attributes constant.  
The unit change is computed with respect to the unprotected group (\texttt{Caucasian} for ethnicity and \texttt{Male} for gender), which is the reference level in the dummy coding. 
Since the $\beta_k$ are not easily interpretable by themselves, we then compute the mean marginal effects of each attribute, i.e., how much the TMR or FMR change when we shift from the unprotected value to a protected one (for instance \texttt{Male} to \texttt{Female}). 
Since we control for all other variables simultaneously, this effect can be interpreted as an effect with all other attributes kept constant. Therefore, the marginal effect estimates the effective demographic biases while accounting for confounding factors.

\autoref{fig_FPR} presents the logit model results for the \origin and \gender attributes on RFW, showing the computed marginal effects on FMR. The marginal effects are calculated relative to each attribute's unprotected reference group. The higher the bar, the higher the bias against the protected subgroup. For example, when using \dcfaceNosp, our analysis shows that the FMR for the \texttt{African} subgroup is 35 points higher than for the \texttt{White} subgroup, independently of the other considered attributes. The addition of re-weighting does not affect this bias, while re-sampling reduces it to 22 points. Our method further reduces it to 12 points. Concerning gender bias, despite decreasing the bias for \originNosp, re-sampling increases the bias for \genderNosp. The proposed controlled generation reduces biases for \origin while keeping the bias in \gender non-significant. The results of the logit model on TMR and on \favcid are provided in the supp. material.

The logit model results provide valuable insights into the fairness implications of different face recognition methods and datasets. By comparing the marginal effects across attributes and methods, we identify the extent and nature of biases of each approach. 
The significantly smaller marginal effects observed in \autoref{fig_FPR} shows our controlled dataset generation reduces biases compared to the original \dcface dataset and baseline mitigation techniques. 
The interpretation of the logit model results highlights the disparities in face recognition performance across different attribute subgroups, showing the importance of considering fairness in the development and evaluation of face recognition systems and the need for effective bias mitigation strategies.

\subsection{ANOVA on Latent Space}
The variation of the performance and fairness metrics across demographic segments can be seen as a consequence of the variability in the distribution of feature vectors in the model's latent space.
Therefore, we utilize ANOVA to investigate the influence of personal attributes on the distances in this latent space.  In our case, the groups are defined by the person's attributes, such as gender, age, and ethnicity, and the explained variable is the distance between face representations in the latent space.
We use the sum of squares computed during ANOVA to extract the $\eta^2$  associated with each attribute. Each $\eta^2$ value represents the impact of the variable on the distance variance in the latent space. The $\eta^2$ of each attribute sum to the $R^2$ of the ANOVA, i.e. the total variance explained by the model.

\begin{figure}[t]
    \centering
    \includegraphics[width=\linewidth]{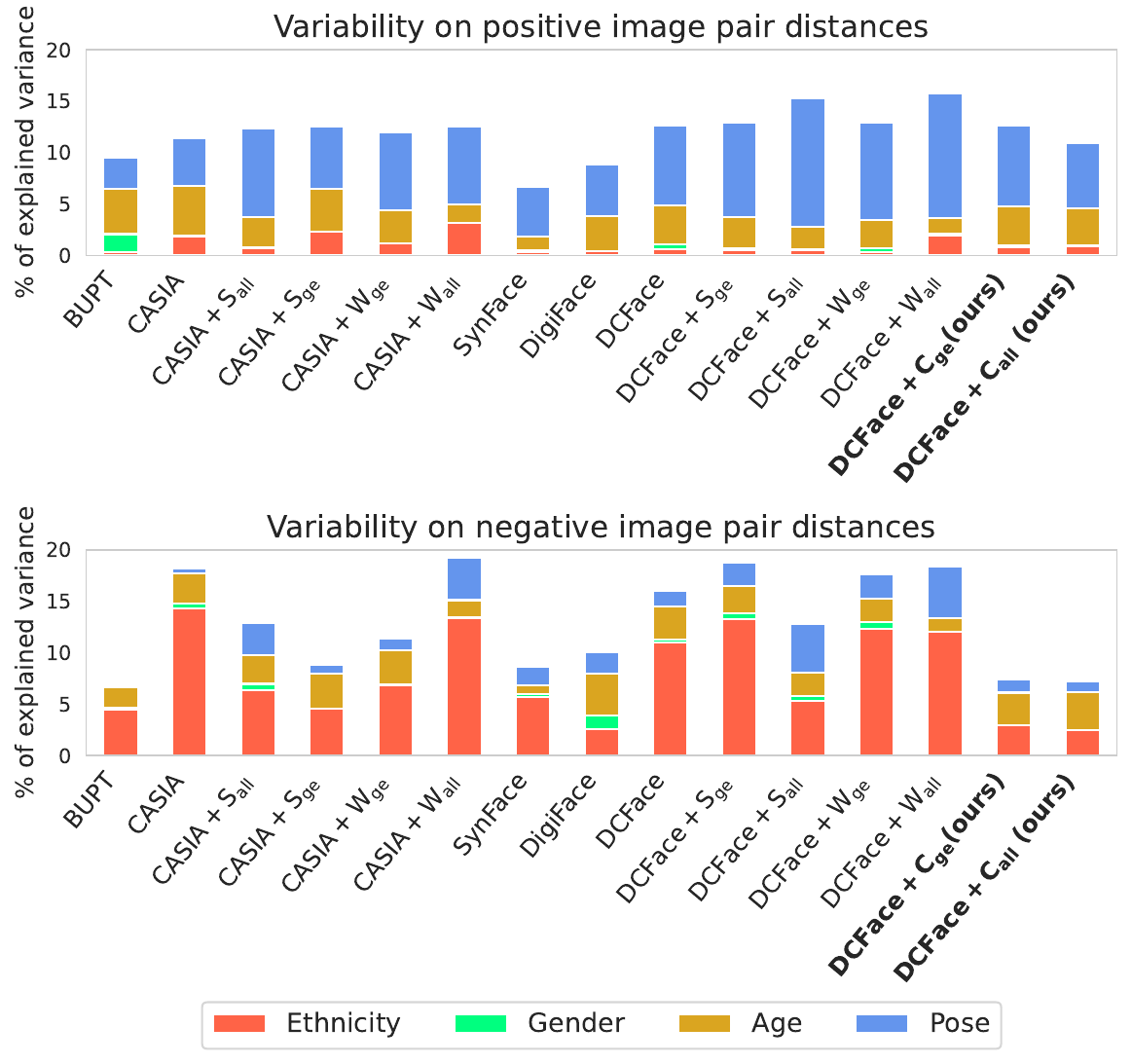}
    \caption{ANOVA results on RFW: total height corresponds to $R^2$, the explained variance by the variables. Each bar is decomposed into multiple $\eta^2$, i.e. the individual contributions to the variance.}
    \label{fig_ANOVA}
\end{figure}
\autoref{fig_ANOVA} shows the result of ANOVA on the distances in the latent space of the RFW dataset, both on the positive and negative pairs. As expected, the explained variance on the positive pairs is generally smaller than the explained variance of the negative pairs, since two images of different people are likely to have more variability than two images of the same person. Moreover, the total $R^2=0.18$ of the ANOVA shows that $18\%$ of variance in the distances in the latent space can be attributed solely to the considered people's attributes.
\pose has the strongest influence on the positive pairs, a finding explained by the strong pose variability for the same person. 
However, neither \origin nor \gender play a big role, meaning that across demographic segments, the spread of the latent vectors of a single person is very similar. 
This is expected since the training loss tries to bring closer the latent vectors of the same individual, who has only one \origin and \gender value.

On the negative pairs, \origin is the attribute having the highest impact on the latent vectors. 
This means that the distances for negative pairs are much higher for some demographic segments than others. 
This result quantifies how much the demographic imbalance translates into the geometry of the latent space. 
Confirming previous works \cite{karkkainen2019fairface, bias1} with another approach, our analysis shows a significant impact of the demographic attributes on the spread of the latent vectors. 
Once more, the impact of the proposed datasets, \dcfaceAll and \dcfaceOGNosp, on the $\eta^2$ shows the effectiveness of our controlled generation. By contrast, traditional training strategies such as re-sampling and loss-weighting are not as good at mitigating the biases in the latent space. 

\section{Limitations}
Attribute inference tools are needed to obtain demographic attributes but introduce prediction errors. 
FairFace is widely used in the field~\cite{kim2023dcface,sarridis2023fair,wang2021meta} and could be improved, particularly for a finer-grained \origin detection. 
The statistical bias analysis is sensitive to the variability and size of evaluation sets.
Consequently, it should only be applied to datasets having sufficient samples for each demographic segment to obtain significant results.
This discards using classical datasets, such as LFW~\cite{LFWTechUpdate}, for fairness analysis.
We present a dataset-balancing pipeline combining several attributes and implement it for DCFace~\cite{kim2023dcface}, a recent and competitive face generator.
The pipeline can be adapted to other generators, such as IDiff-Face~\cite{boutros2023idiff}.
\origin and \gender are the most sensitive attributes, and their balancing focuses on seed ID images needed to drive identity generation in most existing methods.  

\section{Conclusion}
We addressed FVT fairness by evaluating the performance and bias of models trained on various real and synthetic datasets. We proposed a novel controlled generation approach to create balanced synthetic datasets, \dcfaceOG and \dcfaceAllNosp, which prioritize attribute diversity. Our experiments demonstrated that models trained on balanced datasets significantly improved face verification fairness metrics while maintaining competitive accuracy. 
The proposed analysis based on logit regression and ANOVA revealed that the controlled generation method effectively reduces attribute-based biases in both model predictions and latent space representations. It also highlights a persistent disparity in fairness across all considered approaches, which penalizes the African subgroup in particular.  

Our findings have important implications for developing fairer and more inclusive FVT systems. 
By demonstrating the effectiveness of attribute balancing in synthetic data generation and providing a comprehensive evaluation framework, we advocate for more efforts in addressing bias issues in computer vision applications. 
Future research could explore integrating our approach with other bias mitigation techniques and investigate the generalizability of our findings to other computer vision tasks and datasets.

\paragraph*{Acknowledgement} This work was partly supported by the SHARP ANR project ANR-23-PEIA-0008 in the context of the France 2030 program and the STARLIGHT project funded by the European Union’s Horizon 2020 research and innovation program under grant
agreement No 101021797. It was made possible by the use of the FactoryIA supercomputer, financially supported by the Ile-De-France Regional Council.

{\small
\bibliographystyle{ieee_fullname}
\bibliography{egbib}
}

\newpage

\include{supp_t}
\end{document}

%% file: result_table.tex
\begin{table*}[!h]
\centering
\resizebox{\linewidth}{!}{
\begin{tabular}{l|cccccc|cccccc|cccccc}
 & \multicolumn{6}{c|}{RFW~\cite{wang2019racial} } & \multicolumn{6}{c|}{FAVCI2D~\cite{popescu2022face}} & \multicolumn{6}{c}{BFW~\cite{robinson2020face}} \\ \cline{2-19}
 &DoB$\downarrow$ & DPD$\downarrow$ & EOD$\downarrow$& DPR$\uparrow$  & EOR$\uparrow$ & Acc$\uparrow$&DoB$\downarrow$ & DPD$\downarrow$  & EOD$\downarrow$& DPR$\uparrow$ & EOR$\uparrow$ & Acc$\uparrow$&DoB$\downarrow$ &  DPD$\downarrow$ & EOD$\downarrow$& DPR$\uparrow$  & EOR$\uparrow$ & Acc$\uparrow$ \\
\midrule
\midrule
BUPT                & \textbf{30.3} & 23.6 & \textbf{11.9} & 68.4 & 28.6 & \textbf{89.5} & \textbf{38.4} & \textbf{13.0} & \textbf{14.4} & \textbf{75.2} & 19.3 & \textbf{81.8} & \textbf{25.7} & \textbf{5.5}  &\textbf{ 12.5} & \textbf{88.8} & \textbf{}\textbf{26.1} & \textbf{92.6} \\
CASIA               & \underline{35.3} & 19.0 & 22.0 & \underline{71.1} & 5.2  & \underline{85.1} & \underline{39.0} & 21.2 & 28.5 & 66.3 & 16.5 & \underline{81.1} & \underline{29.1} & \underline{9.2}  & \underline{14.9} & \underline{82.2} & 1.3  & \underline{90.3} \\
CASIA + S$_{eg}$    & 39.4 & \textbf{11.5} & \underline{18.8} & \textbf{80.4} & 29.4 & 79.9 & 43.1 & \underline{15.5} & \underline{18.7} & \underline{70.6} & 31.6 & 75.1 & 31.8 & 10.0 & 18.9 & 80.4 & 11.3 & 88.0 \\
CASIA + S$_{all}$   & 48.2 & 17.8 & 24.0 & 68.8 & \textbf{43.7} & 62.3 & 48.6 & 22.0 & 24.1 & 60.3 & \textbf{43.8} & 61.8 & 43.5 & 19.3 & 33.6 & 67.8 & 23.1 & 74.0 \\
CASIA + W$_{eg}$    & 43.5 & \underline{17.3} & 22.8 & 70.2 & 23.8 & 74.0 & 45.3 & 22.3 & 21.0 & 59.2 & \underline{32.7} & 71.1 & 35.4 & 12.5 & 18.5 & 77.0 & 18.5 & 84.9 \\
CASIA + W$_{all}$   & 49.1 & 26.7 & 36.2 & 54.7 & \underline{30.7} & 59.1 & 49.1 & 28.1 & 31.2 & 47.1 & 31.0 & 59.4 & 46.4 & 29.6 & 38.5 & 52.4 & \underline{23.5} & 68.2 \\
\midrule
SynFace             & 48.6 & 13.8 & 24.9 & 73.6 & \underline{44.0} & 60.9 & 48.5 & 22.7 & 26.4 & 57.3 & 37.4 & 62.0 & 45.4 & 20.4 & 23.2 & 63.3 & \underline{36.4} & 70.7 \\
DigiFace            & 45.9 & 15.5 & 25.6 & 73.6 & 37.0 & 69.2 & 47.3 & 21.0 & 22.2 & 62.1 & 40.4 & 66.0 & 45.7 & 16.0 & 21.1 & 70.1 & \textbf{44.8} & 70.0 \\
DCFace              & 42.7 & 17.2 & 32.7 & 71.4 & 15.3 & 75.6 & 45.1 & 20.0 & 18.9 & 62.8 & 32.1 & 71.6 & 35.4 & 14.2 & 21.5 & 74.4 & 11.7 & 85.0 \\
DCFace + S$_{eg}$   & 44.0 & 13.7 & 36.7 & 76.5 & 18.2 & 72.3 & 45.9 & 15.5 & 21.2 & 68.4 & 31.5 & 69.5 & 37.2 & 18.6 & 29.7 & 68.3 & 10.1 & 82.9 \\
DCFace + S$_{all}$  & 48.0 & 16.7 & 23.8 & 69.5 & 38.7 & 63.6 & 48.1 & 22.1 & 23.0 & 58.2 & 43.8 & 63.4 & 42.9 & 16.8 & 25.8 & 68.7 & 21.8 & 75.3 \\
DCFace + W$_{eg}$   & 44.2 & 16.7 & 33.4 & 70.7 & 18.9 & 72.7 & 46.0 & 19.2 & 20.9 & 62.1 & 29.9 & 69.4 & 36.9 & 14.6 & 20.1 & 72.3 & 12.5 & 83.5 \\
DCFace + W$_{all}$  & 49.0 & 19.4 & 31.6 & 59.9 & 23.4 & 59.9 & 48.5 & 23.7 & 26.0 & 54.9 & 36.5 & 61.8 & 44.4 & 24.0 & 24.3 & 56.5 & 27.3 & 72.8 \\
\cdashline{1-19}
\textbf{DCFace + C$_{eg}$}  & \underline{42.2} & \underline{12.7} & \textbf{13.7} & \underline{77.1} & 41.2 & \underline{76.4} & \underline{44.7} & \underline{14.3} & \underline{15.6} & \textbf{71.1} & \textbf{66.0} & \underline{72.4} & \underline{34.7} & \textbf{11.3} & \underline{13.8} & \textbf{77.8} & 23.0 & \underline{85.7} \\
\textbf{DCFace + C$_{all}$} & \textbf{41.6} & \textbf{11.2} & \underline{14.6} & \textbf{80.3} & \textbf{45.9}  & \textbf{77.3} & \textbf{44.5} & \textbf{14.2} & \textbf{14.9} & \underline{70.9}& \underline{58.6} & \textbf{72.7} & \textbf{34.2} & \underline{11.5} & \textbf{13.5} & \underline{77.5} & 24.1 & \textbf{86.1} \\

\end{tabular}
}
    \caption{Fairness metrics and Micro-average accuracy scores of tested datasets and bias mitigation techniques. Real and synthetic datasets are separated. Groups are defined as a combination of \gender and \origin. DPD: Demographic Parity Difference; EOD: Equalized Odds Difference;  DPR: Demographic Parity Ratio; Equalized Odds Ratio; Acc: Micro-average Accuracy. The best results for each dataset type are in \textbf{bold}, and the second-to-best results are \underline{underlined}.}
\vspace{-3mm}
    \label{tab_results}
\end{table*}

%% file: supp_t.tex
\appendix 
\section*{Supplementary Material}
\section{Parameters for training and generation}
For training the face classifier, we use the Adaface training pipeline \cite{kim2022adaface}. We apply the same augmentations, crop, and low-resolution augmentations, for all training sets, with an exception on \digiface, where we also use the augmentation of the authors to reach optimal performances. We perform the training on 4 GPUs with a batch size of 256 (i.e. 64 per GPU),  the optimizer is the standard SGD with a learning rate of 0.1 and a momentum of 0.9. We use as a scheduler a multi-step learning rate decay whose milestones are the epochs 12,20,24 and the decay coefficient is 0.1. The training loss is that of Adaface~\cite{kim2022adaface}.  The margin parameter m is set to 0.4, and the control concentration constant h to 0.333 as recommended by \cite{kim2022adaface}. On each training set, the training lasts 60 epochs.

For generating the DCFACE set and its variants, we use the generation pipeline of \cite{kim2023dcface}. 
We impose the $X_{id}$ image and the $X_{sty}$ to be of the same demographic group as we found that mismatching is likely to induce non-convergence of the resnet50 model when training on the resulting dataset (in particular when mismatching in gender). 
Randomly sampling the style image within the CASIA dataset thus results in a non-decreasing loss of the ResNet network. Within the code of \cite{kim2023dcface}, there is a sampling strategy we haven't tested: combining DDPM images with the closer CASIA faces. 
This approach was and still is, unfortunately, non-usable due to incomplete critical files \footnote{ The provided center\_ir\_101\_adaface\_webface4m\_faces\_webface \_112x112.pth file doesn't have a required "similarity\_df" field.  Also, the dcface\_3x3.ckpt file doesn't seem to store the following property: recognition\_model.center.weight.data }
Moreover, this strategy is not mentioned in the original paper and, since it combines similar CASIA and DDPM faces in a resnet100 latent space, it seems to be in contradiction with what is stated within the ID Image Sampling subsection of \cite{kim2023dcface}.
We thus chose to ignore this strategy, our study being primarily an analysis of fairness and improvement research in this regard.

For all methods, similarly to what the original paper did, we introduce variability within the considered DDPM $X_{id}$ pictures by using a similar $F_{eval}$ model as in \cite{kim2023dcface}. However, one should be aware that the Cosine Similarity Threshold might vary depending on the training of the $F_{eval}$ network. We used the network trained on \cite{zhu2021webface260m} provided by the Adaface Github repository and found 0.6 as an effective threshold to filter similar images. We also get rid of faces wearing glasses with the following solution \cite{Birskus_Glasses_Detector_2024}.
\section {Performance in Accuracy on other sets}

\input{accuracy_table}

In addition to \favcid, BFW, and RFW, we report in \autoref{tab_error} the raw accuracy results on 5 common evaluation sets used in prior work on the FR task ~\cite{bae2023digiface,kim2022adaface,kim2023dcface,qiu2021synface} : (1) Labeled Faces in the Wild (\lfwNosp)~\cite{LFWTechUpdate}, the reference dataset for the task (2) \calfwNosp~\cite{zheng2017calfw}, a version of \lfwNosp with a larger age variability, (3) \cplfwNosp~\cite{zheng2018cplfw}, a version of \lfwNosp with pose variability, (4) \agedbNosp~\cite{moschoglou2017agedb}, a dataset designed for maximizing age variability, and (5) \cfpfpNosp~\cite{sengupta2016cfp} that is designed for pose variability.

Raw accuracy differs from the micro accuracy reported on the paper. Micro accuracy gives the same importance to each demographic segment, whereas raw accuracy performs a simple mean across all images, without any distinction.

\autoref{tab_error} confirms the performance gain of \dcfaceAll over the original generation pipeline: 
The generation pipeline slightly improves accuracy for four of these datasets (+0.12, +0.36, +0.23, and +0.89 for LFW, CFP-FP, CPLFW, and \favcid) and slightly degrades performance for the other two (-1.43 and -0.36 points for Age-DB and CALFW). On the balanced sets, (i.e. RFW and BFW)  the pipeline induces important gains in accuracy (+2.55 for RFW and +4.06 for BFW).

\section{Bias Mitigation techniques details}
We provide implementation details about the baselines, re-sampling, and loss weighting used to compare with our approach.
\subsection{Re-sampling}
\label{subsec:Re-sampling}

Data re-sampling balances class distribution within training data by employing strategies other than the default uniform sampling. These strategies can consist of over-sampling the data from the under-represented classes and/or under-sampling majority classes \cite{ClassImabalanceImpact, Balancing}.

Oversampling \cite{Oversampling1, Oversampling2, Oversampling3, Oversampling4} increases the number of samples by replicating existing data. However, duplicating data by sampling the several times can lead to over-fitting. On tabular data, interpolating techniques such as SMOTE and its variants \cite{SMOTE, SMOTE1, SMOTE2} can be used in order to tackle this overfitting issue. Still, such approaches are not trivial and more costly for non-tabular data such as images. 

Undersampling, on the other hand, consists in the reduction of the majority classes so that their representativity matches the underrepresented classes. \cite{Undersampling1, Undersampling2, Undersampling3}. The main drawback of such an approach is that it results in unused data, which is not an optimal setup.

Here we use Re-Sampling as a baseline for bias mitigation by combining over-sampling and under-sampling. Specifically, for each attribute $a$ with values $a_j$, we count $n_j$, the number of images with value $a_j$. We then assign a weight $w_j = 1/n_j$ to each image sharing value $a_j$. For each image $x_i$, we compute its weight $w_i$ as the product of the weights of all attributes associated with the image. The sampling probability for each image is calculated as $p_i = w_i / \sum_k w_k$. At each beginning of a training epoch, we sample $N$ images according to the probability distribution $\{p_i\}$, where $N$ is the size of the original dataset.

Note that this approach, coupled with the set of random image augmentations used during training, should mitigate to a certain extent the mentioned limitations of both over-sampling and under-sampling.

\subsection{Loss Weighting}
Loss weighting tries to adapt the loss scale depending of the characteristics of the sample. This weighting can be linked to the difficulty of the sample as done implicitly by the Adaface Loss \cite{kim2022adaface}, which can be induced by the class imbalance or in our use case, by the corresponding attributes representativity. A common way to weight the loss is to use the same weights computed in subsection \ref{subsec:Re-sampling}, i.e. using the invert of the frequency/count \cite{weighting1, weighting2, weighting3}. We thus use the same weights $w_i$ for weighting the loss. The weights are normalized batch-wise to ensure the same order of gradient amplitude. The loss of the batch is defined as:
\begin{equation}
    \mathcal{L}(x_1, ..., x_K) = \frac{\sum_{k} w_k \mathcal{L}(x_k)}{\sum_{k} w_k}
\end{equation}
where $\mathcal{L}(x_k)$ is the sample-wise loss for image $x_i$.
\section{Diagnostics on the regressions}
To be valid, a linear regression needs to satisfy a few properties, mainly:
\begin{itemize}
    \item Correct specification: The model is correctly specified, meaning all relevant variables are included, and no irrelevant variables are included.
    \item Normal distribution of errors: While not strictly necessary for estimation, the assumption that errors are normally distributed allows for valid hypothesis testing and the construction of confidence intervals.
    \item Zero conditional mean (exogeneity): The expected value of the error term is zero for any given value of the independent variables. This implies that the independent variables are uncorrelated with the error term.
    \item Homoscedasticity: The variance of the error term is constant across all levels of the independent variables.
\end{itemize}
For a generalized linear model, such as the logit model, these assumptions are not possible to verify strictly due to the non-linearity of the model. Therefore, we use the DHARMa package \cite{DHARMa} in R to run diagnostics on our models and verify the validity of our regressions. DHARMa uses simulation-based residuals. It creates new data from the fitted model and then calculates the empirical cumulative density function for each observation. This approach allows for standardized residual calculation even for non-normal distributions like in logit models.

\begin{figure}
    \centering
    \includegraphics[width=\linewidth]{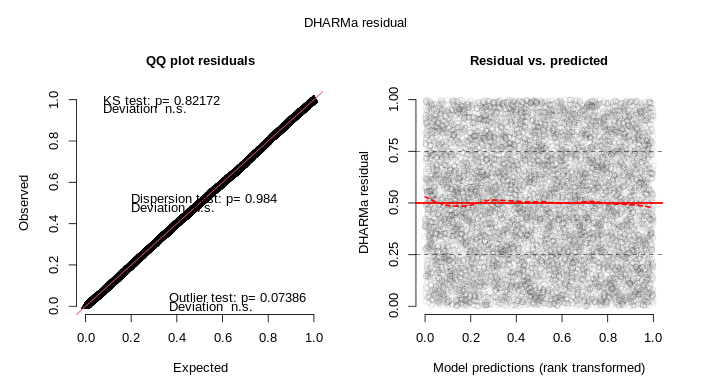}
    \caption{QQ-plot of residuals and Residual vs. predicted plot: logit model is adapted and log-odds are linear in the variables.}
    \label{fig:diag1}
\end{figure}

\begin{figure}
    \centering
    \includegraphics[width=\linewidth]{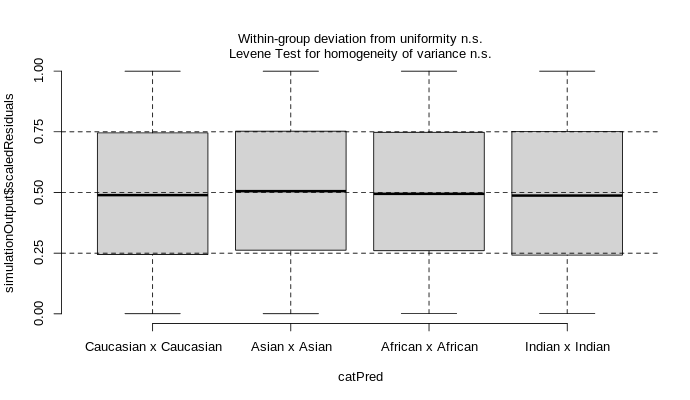}
    \caption{Residual vs. predictor plots: exogeneity is verified.}
    \label{fig:diag5}
\end{figure}

\begin{figure}
    \centering
    \includegraphics[width=\linewidth]{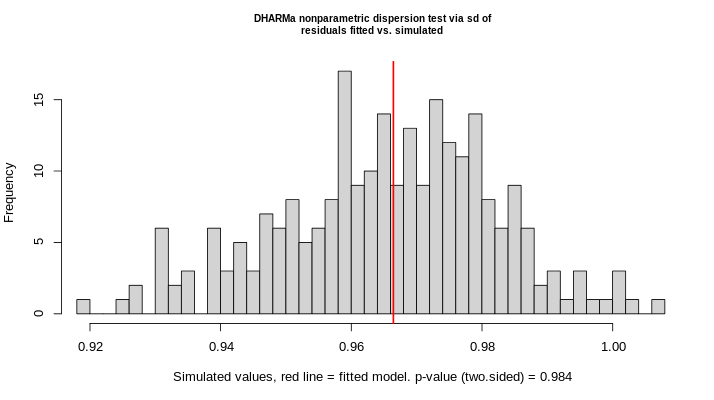}
    \caption{Overdispersion Test: Correct Specification and no auto-correlation.}
    \label{fig:diag3}
\end{figure}

\begin{figure}
    \centering
    \includegraphics[width=\linewidth]{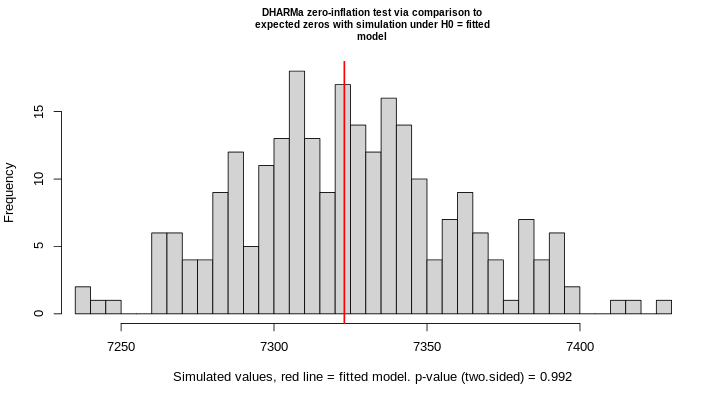}
    \caption{Zero-inflation Test: the model correctly predicts the probability of the outcome.}
    \label{fig:diag2}
\end{figure}

The package provides several diagnostic plots:
\begin{itemize}
    \item QQ-plot of residuals: Checks for overall deviations from the expected distribution (\autoref{fig:diag1}-left).
    \item Residual vs. predicted plot: Helps detect heteroscedasticity and nonlinearity (\autoref{fig:diag1}-right).
    \item Residual vs. predictor plots: Useful for identifying problems with specific predictors (similar to exogeneity) (\autoref{fig:diag5}).
    \item Overdispersion Test: helps to identify if there's more variation in the data than expected under the binomial distribution (\autoref{fig:diag3}).
    \item Zero-inflation Test: check for an excess of zeros or ones (\autoref{fig:diag2}).
\end{itemize}

Here, we will show the diagnostics only for the model \dcfaceAll on RFW, but diagnostics graphs are constant across all tested models on both test datasets.

\section{Statistical Analysis on \favcid}
\begin{figure*}[h]
    \centering
    \includegraphics[width=\linewidth]{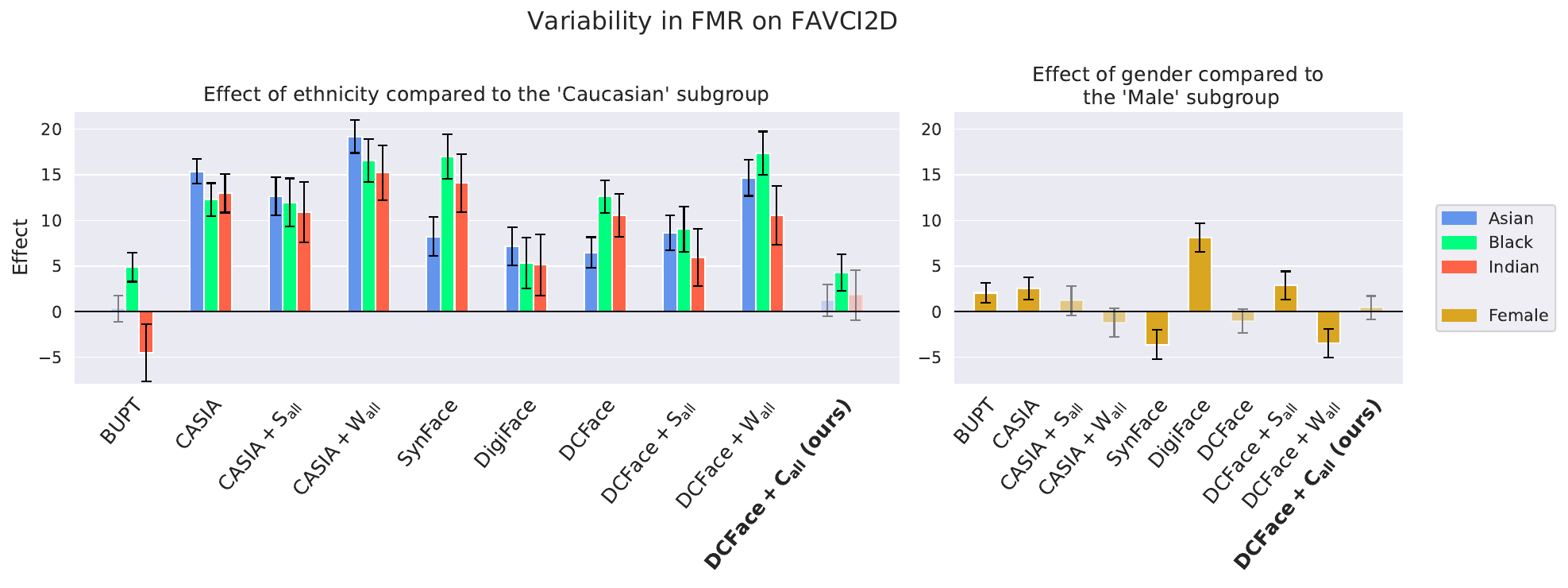}
    \caption{Marginal effect on FMR (lower is better) for each method compared to the unprotected group. Analysis done on \favcid}
    \label{fig:FPRFavcid}
\end{figure*}

\begin{figure*}[h]
    \centering
    \includegraphics[width=\linewidth]{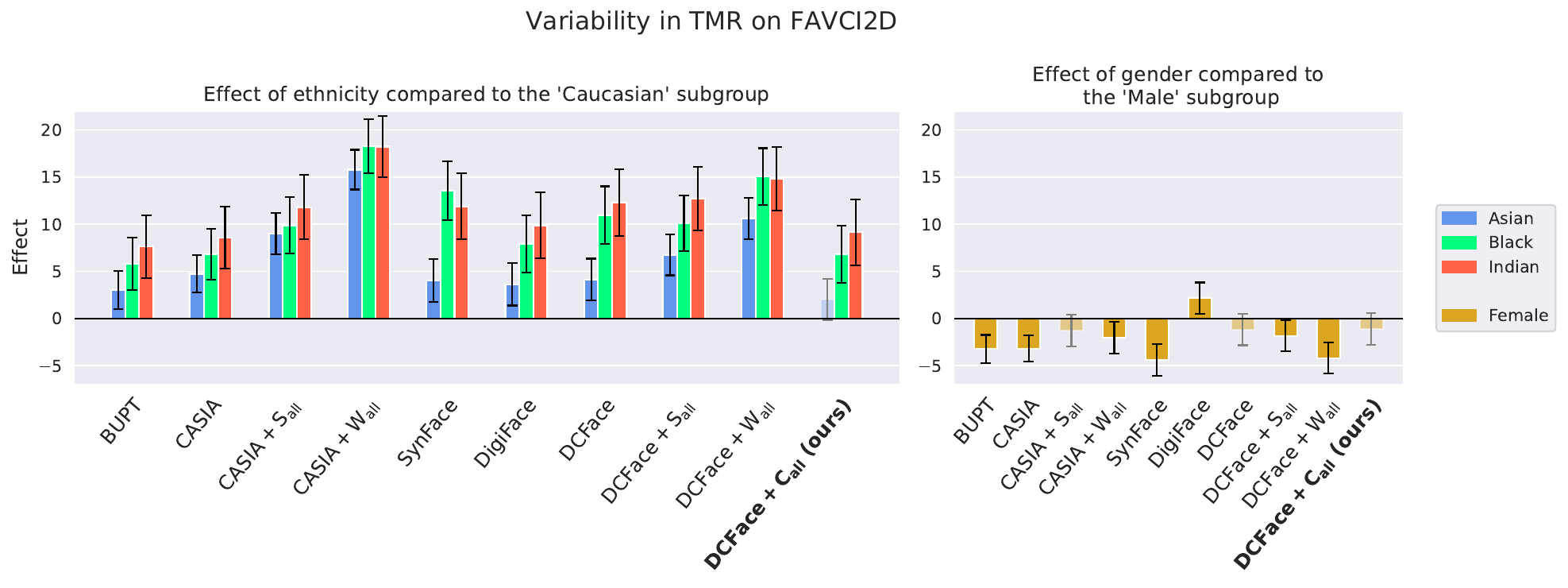}
    \caption{Marginal effect on TMR (lower in absolute is better) for each method compared to the unprotected group. Analysis done on \favcid}
    \label{fig:TPRFavcid}
\end{figure*}

We present here the results of our statistical analysis on \favcid. Be aware that while the metadata of this dataset contains gender information, it doesn't provide ethnicity. We infer it using FairFace. We consider the prediction of FairFace robust enough to compute macro metrics such as the Diversity metric of the main paper however for a finer study such as ours, it might introduce some uncertainty due to model prediction error (\autoref{tab:fairfaceacc}). With that in mind, we still get consistent results for the effects of demographic attributes on the models (\autoref{fig:FPRFavcid}). Our approach shows even more insensitiveness on \favcid than BUPT, by contrast to the results obtained on RFW. The increase of the BUPT-trained model's sensitivity with regard to the inferred labels on \favcid might come from the dataset balancing done on the same labeling system as RFW. Results obtained regarding the TMR (\autoref{fig:TPRFavcid}) and FMR are coherent with the idea that models tend to predict positive outcomes for certain protected ethnical sub-groups. They thus have a high recall for these groups (high TMR and high FMR). With the gender provided by the metadata, we can thus confirm the impact of the balancing on fairness relative to this attribute. While most of the models are sensitive to gender, the model trained on \dcfaceAll \dcface has close to no sensitivity for this attribute, both being close to perfectly balanced concerning gender. 

\begin{figure*}[h]
    \centering
    \includegraphics[width=\linewidth]{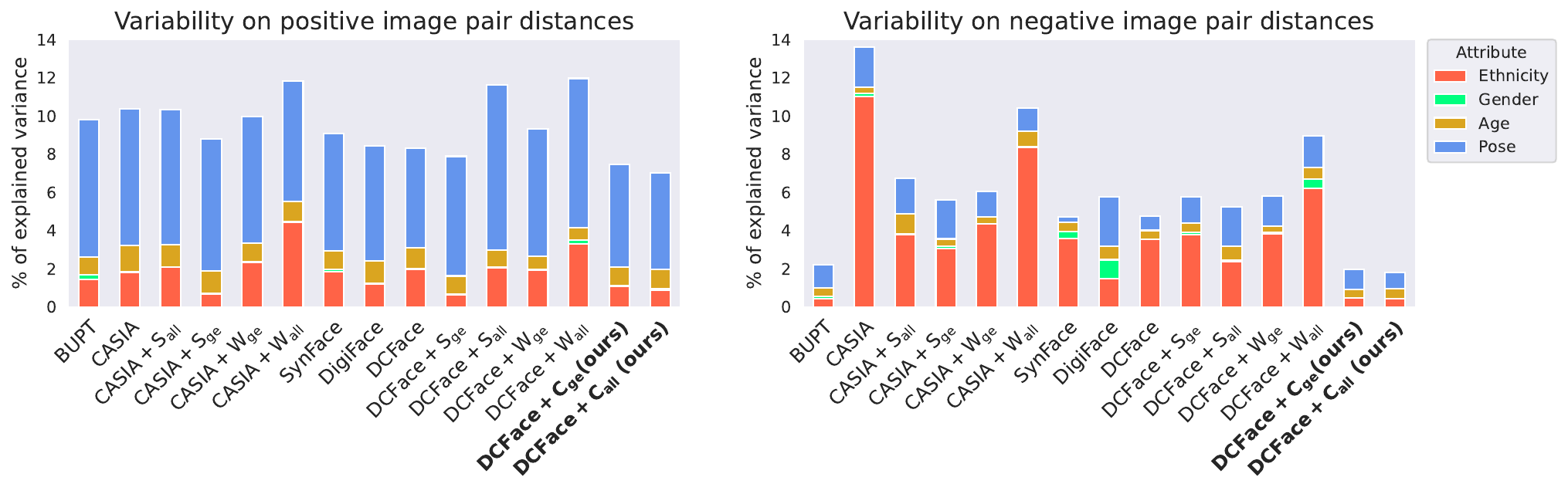}
    \caption{ANOVA results on \favcid: total height corresponds to $R^2$, the explained variance by the variables. Each bar is decomposed into multiple $\eta^2$, i.e. the individual contributions to the variance}
    \label{fig:ANOVAfavcid}
\end{figure*}

\autoref{fig:ANOVAfavcid} shows the result of ANOVA on the distances in the latent space of the \favcid dataset, both on the positive and negative pairs. The results are coherent with the ANOVA computed on RFW. It furthermore highlights the sensitivity of some models' latent space to gender, while our balancing approach allows for more insensitivity about demographic attributes.

\section{Statistical Analysis on BFW}
To tackle the issue of the lack of metadata, in addition to BFW, other alternatives exist such as BFW \cite{bias3} and DemogPairs \cite{demogpairs_hupont_2019}. While these datasets provide some ground-truth metadata, they are composed of significantly fewer identities compared to datasets like \favcid or RFW. This is a limitation of our analysis: Having too few identities might bring instability within Anova or marginal effect studies due to redundancy. We report the results obtained with BFW on as similar number of pairs as RFW and \favcid(24k), meaning every single identity appears in around 30 evaluated pairs. The impact of the number of identities within benchmarking should be studied in future works as this might affect the obtained analysis of performance and fairness.

\begin{figure*}[h]
    \centering
    \includegraphics[width=\linewidth]{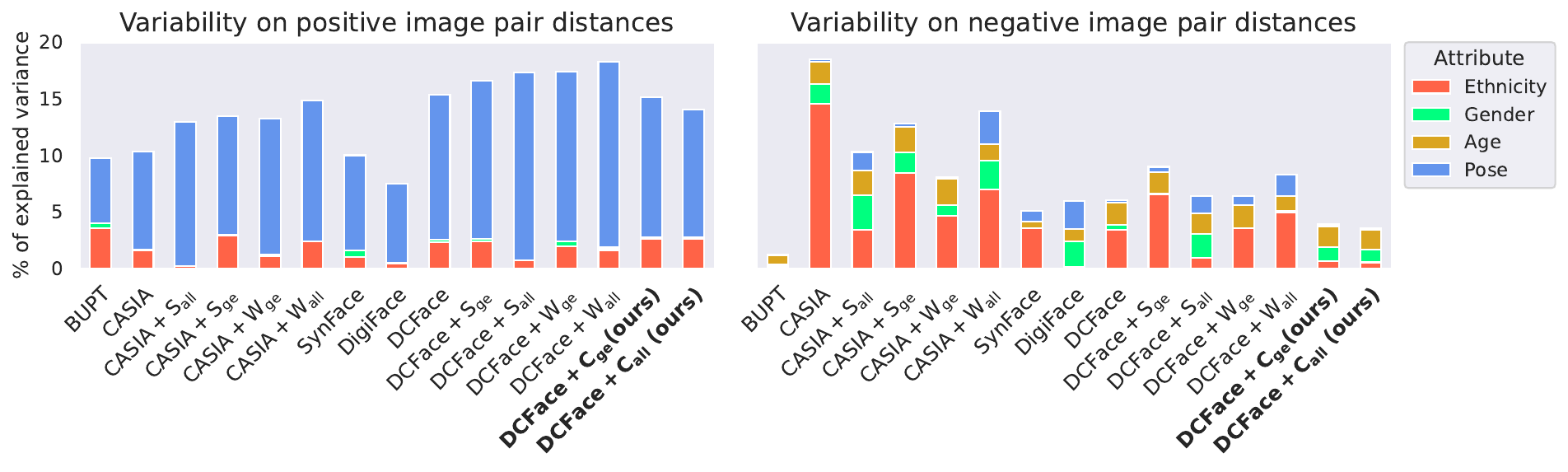}
    \caption{ANOVA results on BFW: total height corresponds to $R^2$, the explained variance by the variables. Each bar is decomposed into multiple $\eta^2$, i.e. the individual contributions to the variance}
    \label{fig:ANOVA_BFW}
\end{figure*}

Figure \ref{fig:ANOVA_BFW} shows the ANOVA analysis performed on BFW. As before, on the negative image pairs, our conditional generation methods greatly reduces the variance explained by the sensitive attributes. 

\begin{figure*}[h]
    \centering
    \includegraphics[width=\linewidth]{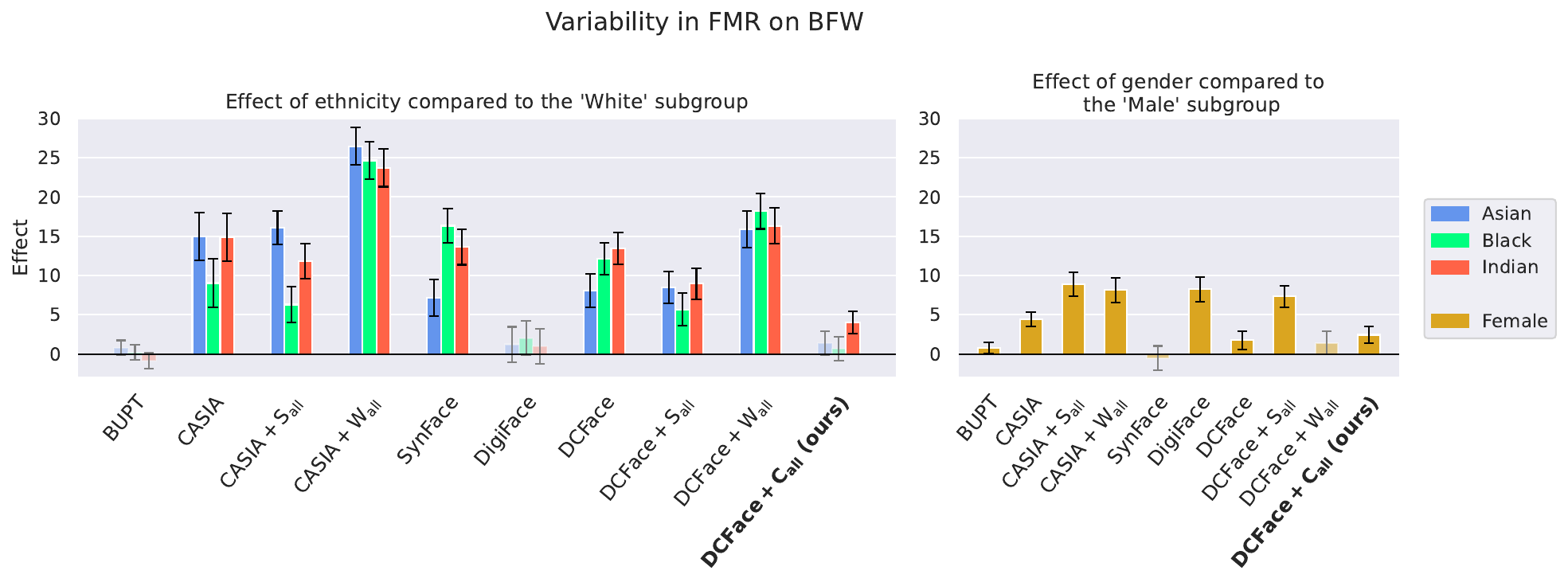}
    \caption{Marginal effect on FMR (lower is better) for each method compared to the unprotected group. Analysis done on BFW}
    \label{fig:FPRBFW}
\end{figure*}

\begin{figure*}[h]
    \centering
    \includegraphics[width=\linewidth]{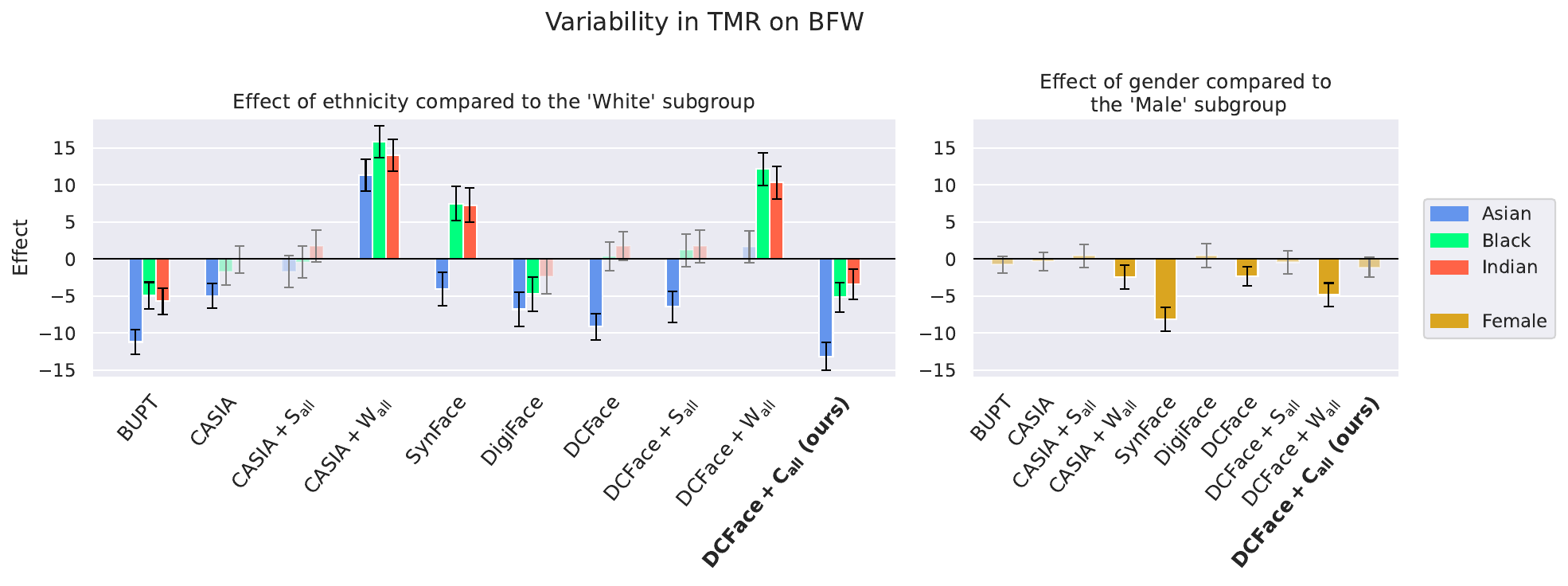}
    \caption{Marginal effect on TMR (lower in absolute is better) for each method compared to the unprotected group. Analysis done on BFW}
    \label{fig:TPRBFW}
\end{figure*}

Figures \ref{fig:TPRBFW} and \ref{fig:FPRBFW} present the marginal effects of the attributes, respectively, on TMR and FMR. As we see, the fairness gain mostly comes from a fairer FMR between ethnicities: the FMR of the Asian and Black subgroups are 8 and 12 points higher than for the White subgroup in the original \dcface, and become non-significant with \dcfaceAll. For the TMR, however, just as for RFW, becomes slightly more unfair between ethnicities. Still, as shown in Table 2 of the paper, on all fairness metrics except EOR, our method outperforms the other synthetic data approaches on BFW.

\begin{table*}
    \centering
    \begin{tabular}{c|ccccccc}
        ethnicity & Black & White & East-Asian & Indian & Latino-Hispanic  & Middle-Eastern & South-Asian \\
        \midrule
        Prediction accuracy & 0.863 & 0.777 & 0.784 & 0.724 & 0.581 & 0.631 & 0.641 \\
    \end{tabular}
    \caption{FairFace model accuracy when inferring on the Fairface validation set. Available Metadata only provides the race7 variable ground truth while we are considering the race variable (whose values are White, Black, Asian, and Indian). The robustness of the model for this latter should be thus greater.}
    \label{tab:fairfaceacc}
\end{table*}

\begin{figure*}[h]
    \centering
    \includegraphics[width=\linewidth]{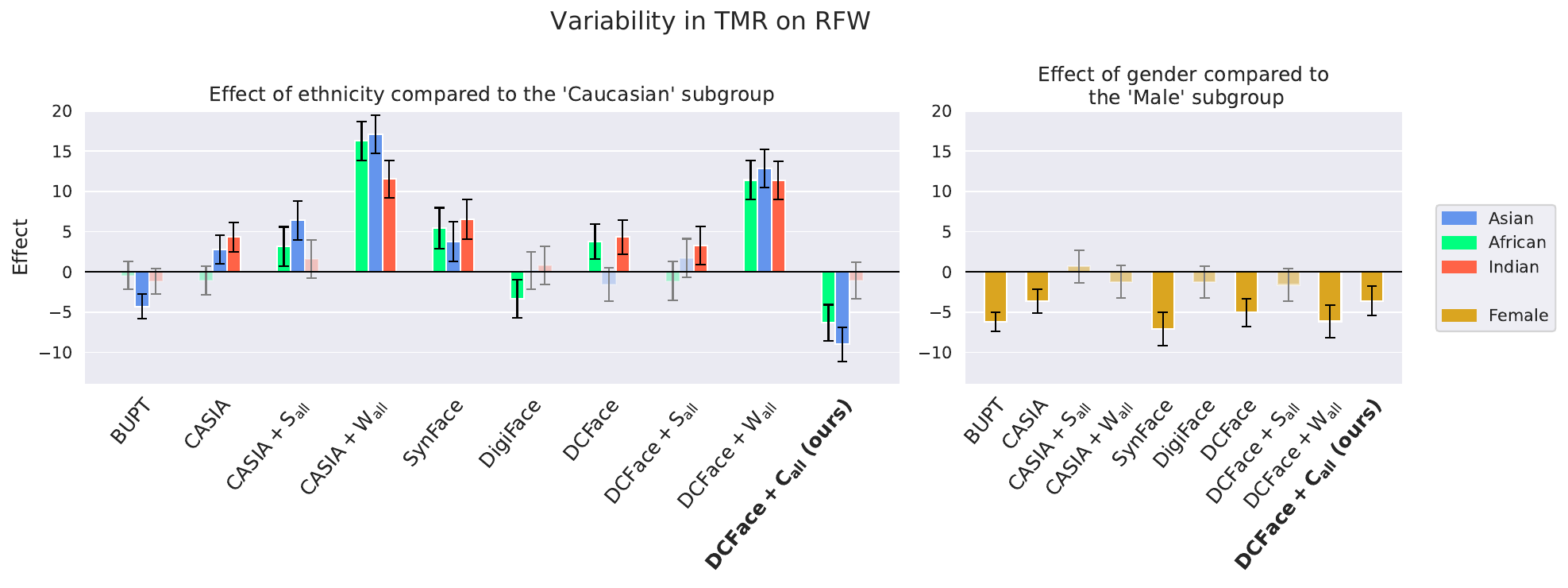}
    \caption{Marginal effects on TMR (lower in absolute is better) for each method compared to the unprotected group. Analysis done on RFW}
    \label{fig:TPRFW}
\end{figure*}
{
\clearpage
\onecolumn
\section{Datasets Images examples}
    \begin{figure}[H]
        \begin{subfigure}[t]{0.45\textwidth}
            \includegraphics[width=\linewidth]{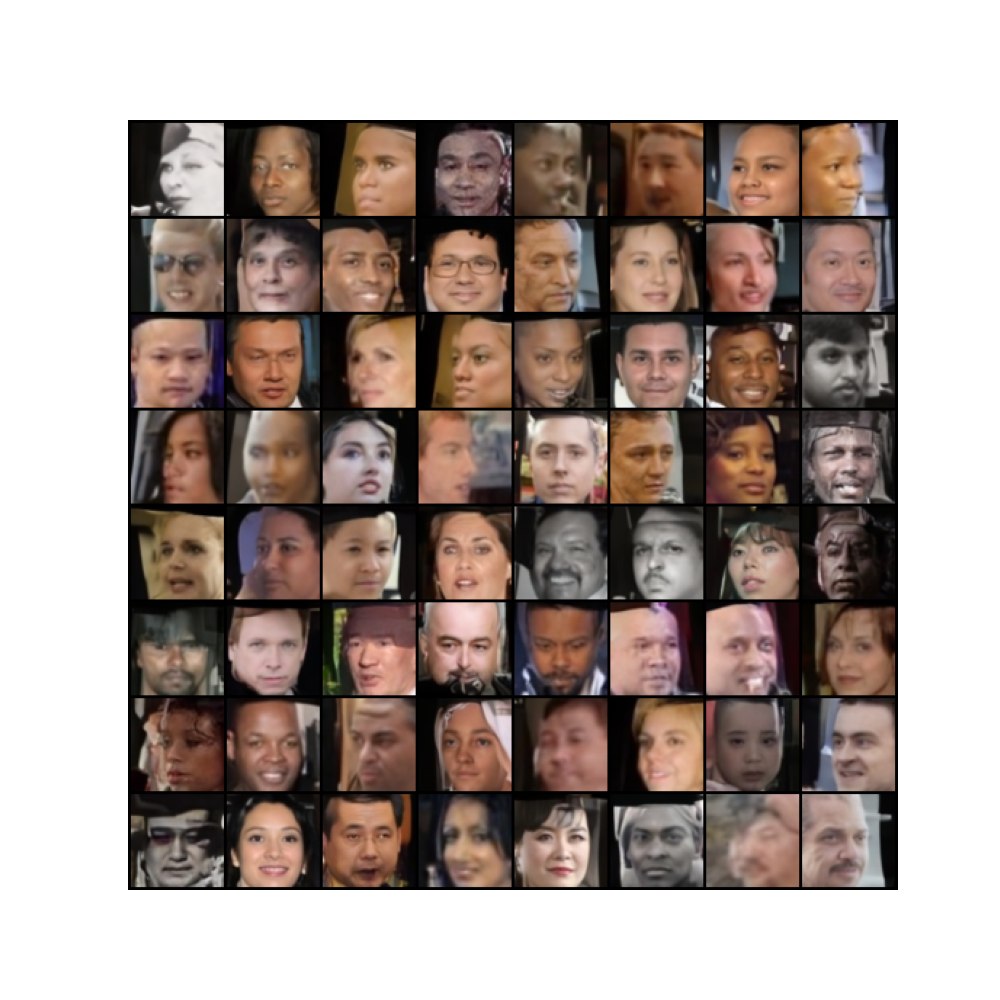}
            \caption{Examples of images within our proposed \dcfaceAll approach. We notice a greater diversity of images.}
            \label{fig:all}
        \end{subfigure}
        \hfill
        \begin{subfigure}[t]{0.45\textwidth}
            \includegraphics[width=\linewidth]{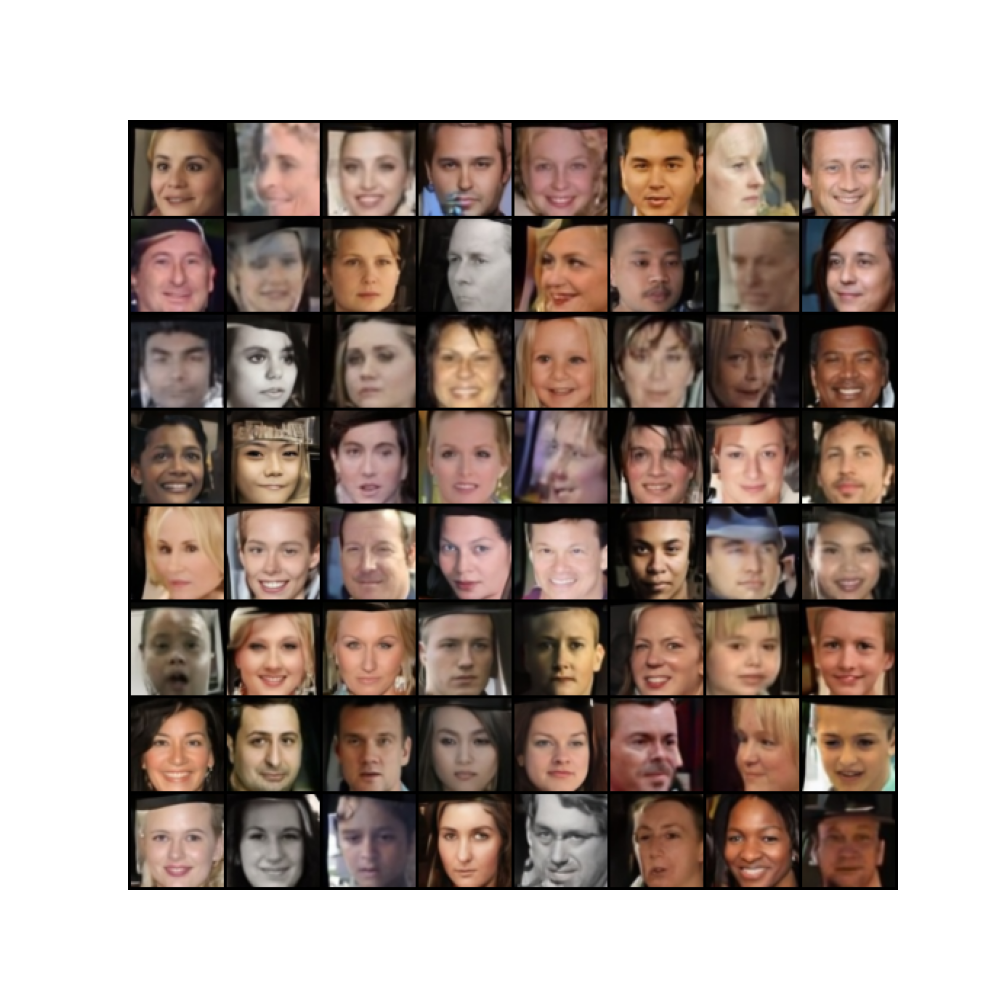}
            \caption{Examples of images generated with the original \dcface \cite{kim2023dcface} pipeline}
            \label{fig:base}
        \end{subfigure}
        \vskip\baselineskip
        \begin{subfigure}[t]{0.45\textwidth}
            \includegraphics[width=\linewidth]{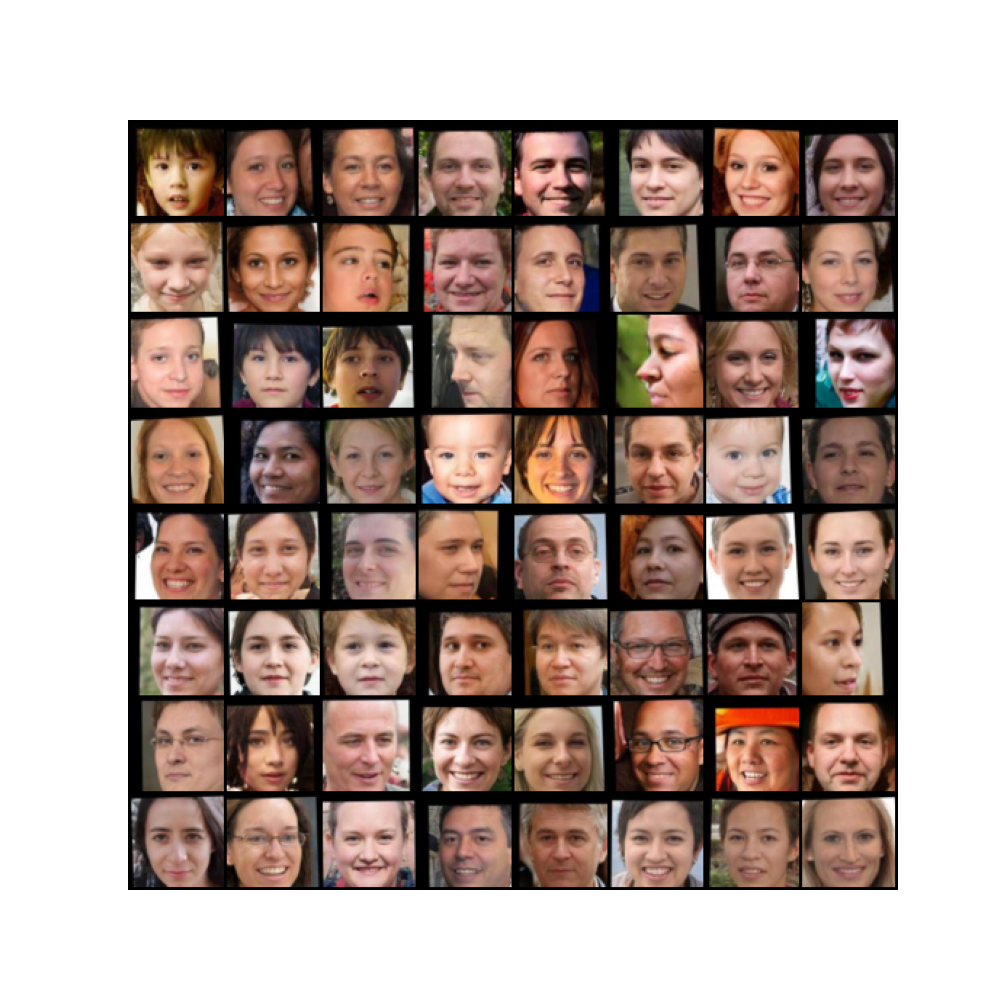}
            \caption{Examples of images generated with the SynFace pipeline \cite{qiu2021synface}}
            \label{fig:synface}
        \end{subfigure}
        \hfill
        \begin{subfigure}[t]{0.45\textwidth}
            \includegraphics[width=\linewidth]{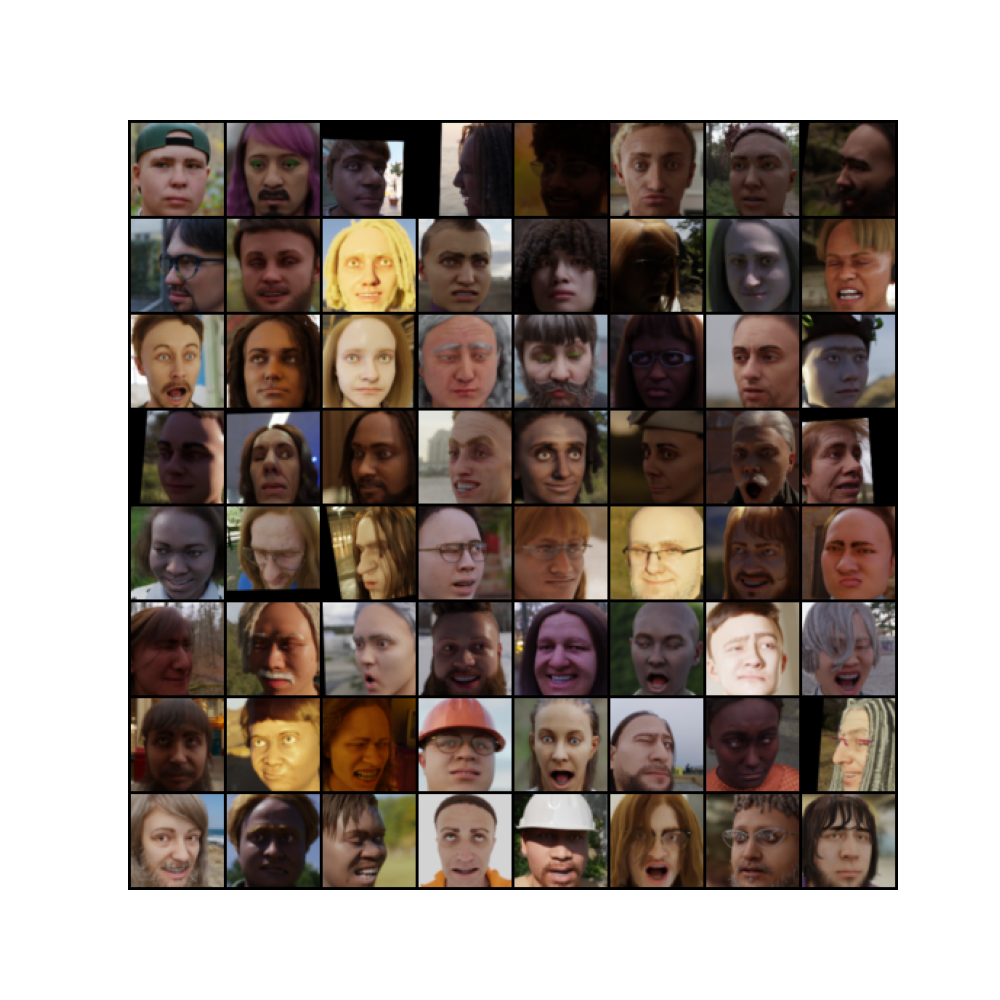}
            \caption{Examples of images within the \digiface dataset \cite{bae2023digiface}}
            \label{fig:digiface}
        \end{subfigure}
    \end{figure}
\FloatBarrier

\begin{figure}[H]   
    \begin{subfigure}[t]{0.48\textwidth}
    \setcounter{subfigure}{4}
        \centering
        \includegraphics[width=\linewidth]{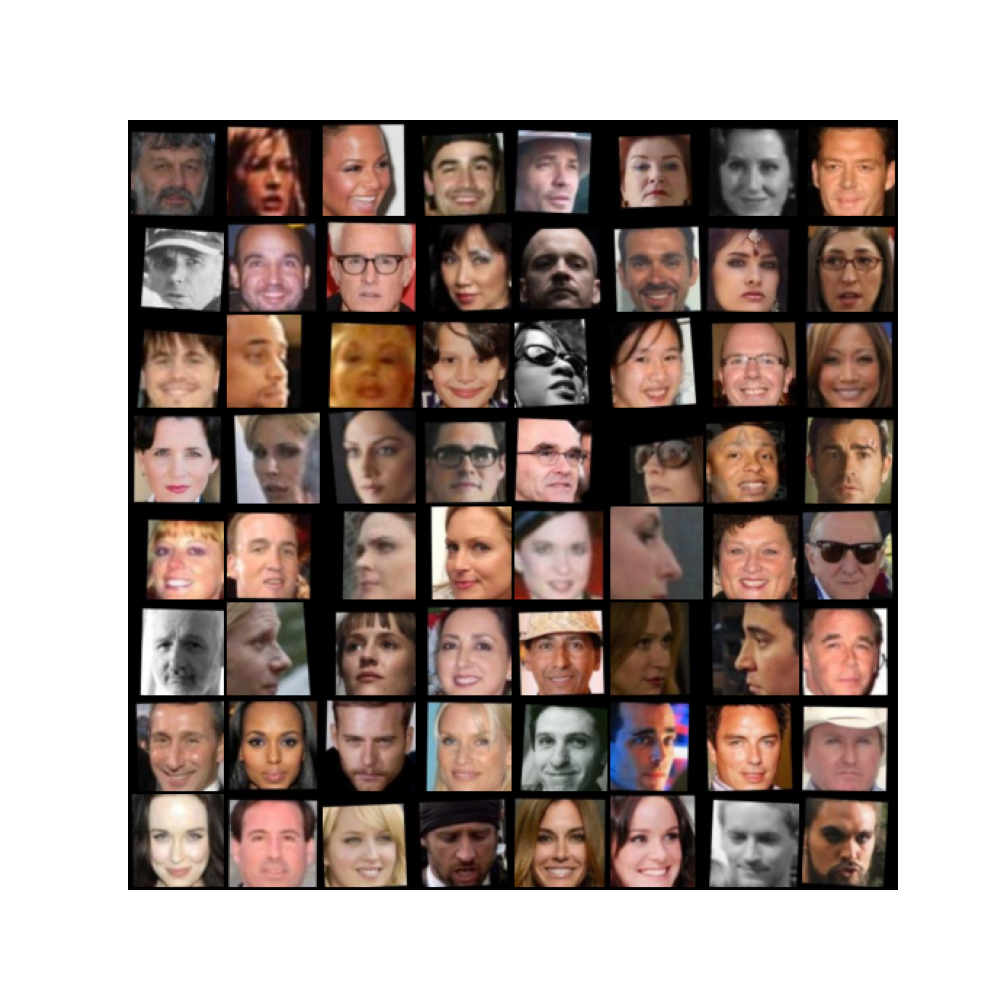}
        \caption{Examples of images within the \casia dataset \cite{yi2014learning}}
    \end{subfigure}
    \hfill
    \begin{subfigure}[t]{0.48\textwidth}
        \centering
        \includegraphics[width=\linewidth]{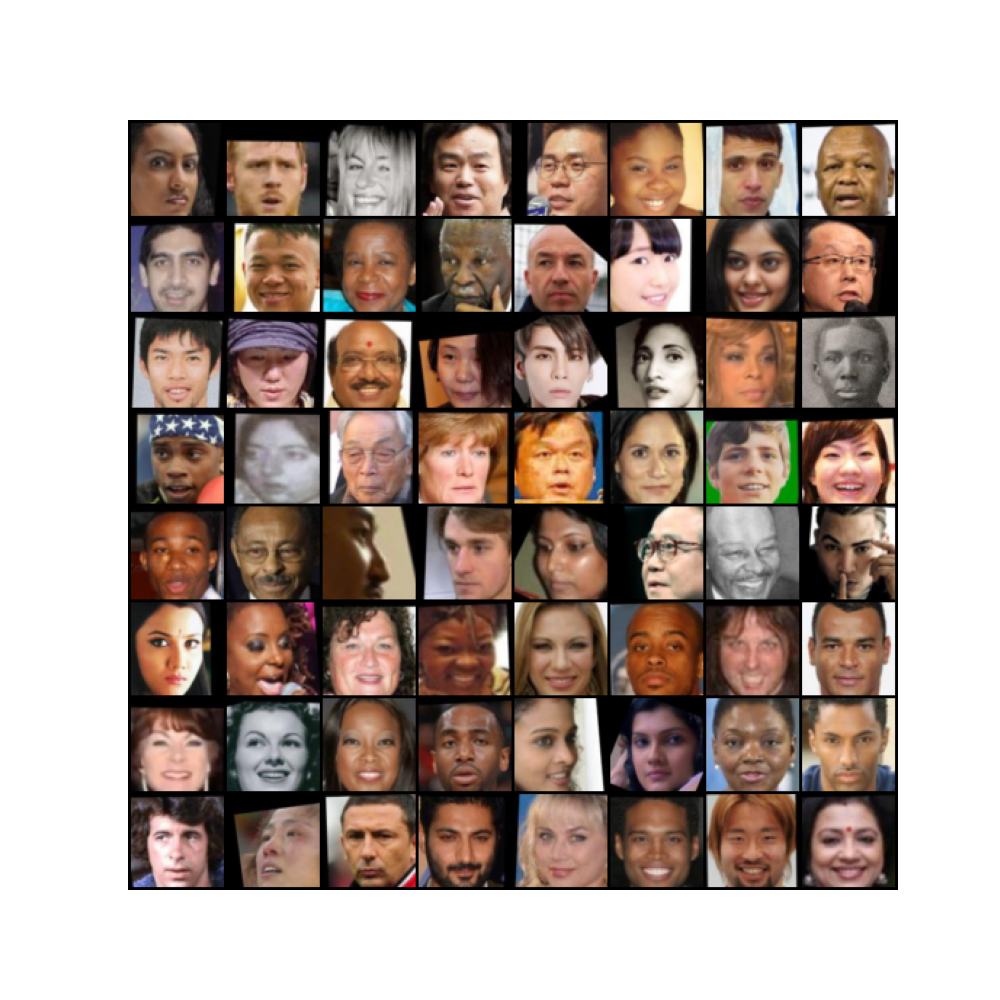}
        \caption{Examples of images within the BUPT dataset \cite{wang2021meta}}
    \end{subfigure}
    
\end{figure}
\FloatBarrier
}

%% file: accuracy_table.tex
\begin{table}[!h]
\begin{center}
    
\resizebox{0.99\linewidth}{!}
{
\begin{tabular}{c|c|c|c|c|c|c|c}

\multirow{2}{*}{\makecell{Verif.\\ dataset}} & \multicolumn{2}{c}{Real dataset} & \multicolumn{5}{|c}{Synthetic datasets} \\ 
\cline{2-8}
& \casia & BUPT & SynFace & \digiface & \dcface & \dcfaceOGNosp & \dcfaceAllNosp \\
\hline
\lfwNosp & 99.46 & \textbf{99.55} & 87.28 & 94.88 & 98.13 & 98.24  & \textbf{98.25}\\
\hline
\cfpfpNosp & \textbf{94.87} & 90.03 &67.01 & \textbf{83.4} & 80.92 & 80.03 &  81.28\\
\hline
\cplfwNosp & \textbf{90.35} & 85.98 &64.91 & 76.61 & 79.94 & 79.32 &  \textbf{80.17}\\
\hline
\agedbNosp & \textbf{94.95} & 94.3 &61.78 & 78.26 & \textbf{87.96} & 86.77 &  86.53\\
\hline
\calfwNosp & 93.78 &\textbf{94.38} &73.53 & 79.78 & 90.39 & \textbf{90.6} & 90.03\\
\hline 
RFW & 86.38 & \textbf{90.35} &64.3 & 72.73 & 76.95 & 78.51 & \textbf{79.5}\\
\hline 
\favcid & \textbf{82.77} & 81.81 & 61.19 &  67.17 & 72.84 & 73.31 &  \textbf{73.73}\\
\hline 
BFW & 89.3 & \textbf{92.48} &70.08 & 77.27 & 84.47 & 85.45 &  \textbf{88.53}\\
\specialrule{.2em}{.1em}{.1em}

AVG & \textbf{91.48} & 91.11 & 68.76 & 78.76 & 83.95 & 84.03 &   \textbf{84.75} \\

\end{tabular}
}
\caption{Raw Accuracy obtained for the different used sets on 8 datasets including five commonly used datasets in addition to BFW, RFW and \favcid}
\label{tab_error}
\end{center}

\end{table}